\documentclass{article} 
\pdfoutput=1  
\usepackage{fullpage}
\usepackage[hyphens]{url}  
\usepackage{graphicx} 
\usepackage[numbers]{natbib}  
\usepackage{caption} 
\usepackage{tikz}

\usepackage{color} 

\title{Learning Robust Algorithms for Online Allocation Problems Using Adversarial Training}

\author{Goran Zuzic\footnote{The first two authors contributed equally.}\footnotemark[1]~\thanks{
    (goran.zuzic@inf.ethz.edu)
    ETH Z\"{u}rich
    Switzerland. Part of this work was done when interning at Google Research in Mountain View.
  }
  \and Di Wang\footnotemark[1]~\thanks{
    (wadi@google.com)
    Google Research,
    Mountain View
  }
  \and Aranyak Mehta\thanks{
    (aranyak@google.com)
    Google Research,
    Mountain View
  }
  \and D. Sivakumar\thanks{
    (siva@google.com)
    Google Research,
    Mountain View    
  }
}

\usepackage{amsmath,amsfonts,bm}

\def\eqref#1{equation~\ref{#1}}

\def\1{\bm{1}}

\def\vp{{\bm{p}}}
\def\vq{{\bm{q}}}

\DeclareMathAlphabet{\mathsfit}{\encodingdefault}{\sfdefault}{m}{sl}
\SetMathAlphabet{\mathsfit}{bold}{\encodingdefault}{\sfdefault}{bx}{n}

\DeclareMathOperator*{\argmin}{arg\,min}

\usepackage{cleveref}

\usepackage[utf8]{inputenc} 
\usepackage{url}            
\usepackage{booktabs}       
\usepackage{amsfonts}       
\usepackage{nicefrac}       
\usepackage{microtype}      

\usepackage{wrapfig}
\usepackage{caption}
\usepackage{subcaption}

\usepackage{float}
\usepackage{algorithm}
\usepackage{algorithmic}
\usepackage{graphicx}
\usepackage{subcaption}
\usepackage{xspace}

\usepackage[font=small,labelfont=bf]{caption}

\newcommand{\ALG}{\ensuremath{\text{\normalfont\scshape alg\ }}} 
\newcommand{\ADV}{\ensuremath{\text{\normalfont\scshape adv\ }}} 
\newcommand{\CR}{\ensuremath{\mathrm{CR}}\xspace}

\newcommand{\mc}[1]{\mathcal{#1}}
\newcommand{\nnoise}{\ensuremath{n_{\text{noise}}}\xspace}
\newcommand{\nbatch}{\ensuremath{n_{\text{batch}}}\xspace}
\newcommand{\Trestart}{\ensuremath{T_{\text{restart}}}\xspace}

\newcommand{\Talg}{\ensuremath{T_{\text{alg}}}\xspace}
\newcommand{\Tadv}{\ensuremath{T_{\text{adv}}}\xspace}
\newcommand{\Tadd}{\ensuremath{T_{\text{add}}}\xspace}

\newtheorem{problem}{Problem}

\def\wadi#1{\textcolor{red}{}}
\newcommand{\YL}{YaoLearner}

\begin{document}

\maketitle

\usetikzlibrary{arrows.meta}

\begin{abstract}
We address the challenge of finding algorithms for online allocation (i.e. bipartite matching) using a machine learning approach. In this paper, we focus on the AdWords problem, which is a classical online budgeted matching problem of both theoretical and practical significance. \wadi{Moved AdWords to the front.} In contrast to existing work, our goal is to accomplish algorithm design {\em tabula rasa}, i.e., without any human-provided insights or expert-tuned training data beyond specifying the objective and constraints of the optimization problem. \wadi{Do we want to mention anything regarding this approach breaks the fundamental limits of existing work and makes it viable to generalize to new problems etc.?}

We construct a framework based on insights and ideas from game theory, adversarial training and GANs \wadi{Do we want to mention GANs? Reviewers may raise it anyway if we don't}. Key to our approach is to generate adversarial examples that expose the weakness of any given algorithm. A unique challenge in our context is to generate complete examples from scratch rather than perturbing given examples and we demonstrate this can be accomplished for the Adwords problem.

We use this framework to co-train an algorithm network and an adversarial network against each other until they converge to an equilibrium. This approach finds algorithms and adversarial examples that are consistent with known optimal results. Secondly, we address the question of robustness of the algorithm, namely can we design algorithms that are both strong under practical distributions, as well as exhibit robust performance against adversarial instances. To accomplish this, we train algorithm networks using a mixture of adversarial and practical distributions like power-laws; the resulting networks exhibit a smooth trade-off between the two input regimes.

\end{abstract}


\section{Introduction}
A long-standing ambition of machine learning is to learn from \emph{first principles}, that is, an ML system that learns {\em without any human-provided data\/} beyond an appropriate definition of the task at hand~\citep{silver2018general}. There have been impressive strides towards this goal, most notably in the domain of tabletop games such as chess, Go, shogi~\citep{silver2018general} and poker~\citep{brown2018superhuman}, where programs have been trained to play at superhuman levels without consuming any game-specific human expertise except, naturally, the rules of the game.

Our work falls under the broad umbrella of using ML to {\em design algorithms for optimization problems\/}. Similar to tabletop games, algorithm design has succinct and well-defined objectives, so learning from first principles is plausible.
%
%
Conceptually, our goals are two-fold:
From a theoretical perspective, we would like to train online algorithms that are optimal in the {\em worst case\/}, as traditionally done in algorithm design and analysis. We aim to achieve this without any predetermined training data (\emph{tabula rasa}).
The second goal of our work is to design optimization algorithms that work well under distributions of inputs that arise in practice, while being \emph{robust} to (possibly adversarial) ``shocks,'' that is, sharp changes in the input distribution. We will refer to this as the {\em robust stochastic} setting. 

In this work we focus on a concrete problem, \emph{AdWords}, which is a classic and complex online problem generalizing both the problems of online bipartite matching and online vertex-weighted bipartite matching. We note that AdWords (and online bipartite matching in general) has been extensively studied, and in the worst case setting the optimal algorithms as well as instances that certifies the hardness of the problem are both well understood. However, the significance of our result lies on the fact that we learn the algorithm without any human intervention in coming up with a training set and none of the expert knowledge is involved in training. We only apply the expert knowledge to validate that the learned algorithm is very close to the theoretically known optimal both qualitatively and quantitatively, and the self-generated hard instances indeed discover the essential hardness combinatorial structure of the problem. It is exactly this tabula rasa nature of our framework that distinguishes our result from (as far as we know) all prior work in the context of learning worst-case algorithms and makes our framework potentially interesting for learning algorithms of less well-understood problems.

\textbf{Overview of Techniques.}
We utilize \emph{adversarial training}~\cite{GSS14,madry2018} as our main technique. In the canonical setup of adversarial training, an adversary adds bounded perturbations to training instances to exploit the weakness of the algorithm and thus making it more robust. When learning worst-case algorithms, We essentially work in the extreme case where arbitrary perturbation is allowed, which is equivalent to constructing completely new instances, to get the most robust algorithm (i.e., worst-case optimal). Thus we need to design a more advanced adversary to tackle this challenge. 

In contrast to most prior work that considers adversarial training through the lens of robust optimization, we draw insights from game theory and online learning (in particular,~\cite{FS96}).
The game-theoretic viewpoint of algorithm design (see Appendix~\ref{sec:game-theory} for details) defines a zero-sum game between an algorithm player and an adversary player, where the optimal algorithm is the max-min strategy of the former. At a high-level, our technique learns an algorithm iteratively: we consider the current version of the algorithm, find its worst-case input (also called ``adversarial input'', since one can imagine a hostile adversary hand-picking it as a best response in the game), and change the algorithm to improve its performance on this input. Naturally, finding the worst-case input is often a hard problem in itself. We propose a framework that uses deep learning both to guide our search for worst-case inputs (using an adversary network) and to improve the algorithm (represented by an algorithm network). We will also show that the framework can be adapted to solve the problem of finding algorithms for the robust stochastic setting. 

We note that the generic structure of a competitive game with two players implemented using neural networks is also reminiscent of GANs~\cite{GAN}. We point out that in spite of this high-level similarity, our framework is quite different from GANs in many ways. For example, the top level goal of a GAN is to construct a generator that mimics some fixed target data distribution, and it’s undesirable if the generator only finds best responses to the discriminator (i.e. mode collapsing) and much of the effort is to avoid this. On the contrary, our framework draws insight from no-regret dynamics (see Appendix \ref{sec:game-theory} for details), where convergence crucially relies on the adversary to quickly collapse to the best response to the algorithm, and thus most of our effort is to the exact opposite of GANs. Moreover, in our case the training eventually converges to an equilibrium between the two players, whereas in GANs the objective is mostly hinged upon the external fixed target distribution. For example, in GANs if we have a perfect generator, then it suffices to use a trivial discriminator that just flips coins. However, in our case, even if we start with an optimal algorithm, it’s still difficult to train the adversary to generate strong instances that push the optimal algorithm to its worst-case bound, and vice versa.

\textbf{Summary of results.} 
The main contributions of this paper are two-fold: a concrete realization of the ideas outlined above in the form of a training framework (Algorithm~\ref{alg:yaogan} in Section~\ref{sec:yaogan}), and employing the framework to derive algorithms for the AdWords problem, a complex online optimization problem that generalizes the classic bipartite matching problem as well as the knapsack problem.

The \emph{AdWords} problem~\citep{MSVV07} is a central problem in the area of online advertising.  In this problem, there are $n$ advertisers with budgets $B = (B_0, \ldots, B_{n-1})$, and $m$ ad slots. Each ad slot $j$ arrives sequentially along with a vector $v^j = (v^j_0, v^j_2, \ldots, v^j_{n-1})$ where $v^j_{i}$ is the bid of advertiser $i$ for slot $j$. Once a slot arrives the algorithm must irrevocably allocate it to an advertiser with sufficient remaining budget $r^j_i$, or to no one at all. If slot $j$ is allocated to advertiser $i$, the algorithm collects a revenue equal to $\min(r^j_i, v^j_i)$.
The remaining budget $r_i^j$ of advertiser $i$ is reduced accordingly: initially, $r_i^0 = B_i$ and then $r_i^{j+1} = r_i^j - \min(r^j_i, v^j_i)$ if slot $j$ is assigned to $i$ and $r_i^{j+1} =r_i^j$ otherwise. The objective of the algorithm is to maximize the total revenue over the entire sequence; due to the online nature of the problem, the allocation decisions have to be made without knowledge of the remaining sequence. 
The primary measure of the performance of online algorithms is the \emph{competitive ratio} (CR). For a fixed input instance this is the ratio of the (expected) objective value the online algorithm achieves for that instance to the \emph{offline optimal value} (the best possible objective value for the instance, knowing the entire sequence upfront). The competitive ratio of the algorithm is defined as its worst competitive ratio over all possible inputs. In this paper we are interested in the regime for AdWords when the bids are small relative to the budgets, where the \emph{MSVV algorithm} of \citep{MSVV07} achieves a CR of $1 - 1/e$ (and this is optimal among all online algorithms).

There are two compelling reasons why we study the AdWords problem.  From a conceptual perspective, its algorithmic solution \citep{MSVV07} involves subtle ideas that trade-off greedily picking high-value bids while carefully hedging on future bids by balancing the budgets spent; also, the ideas in establishing the optimality of the algorithm use carefully crafted worst-case inputs. Thus it is an archetypal example of a problem where it is of interest to know whether deep learning methods can step in for both these tasks. Secondly, from a practical perspective, we wish to learn policies that perform better than the worst-case guarantee on the input distributions that arise in practice; at the same time, since traffic patterns can change abruptly based on external events, and advertisers can change their bidding and targeting at any time, it is vital to derive algorithms that are robust to such shocks.
Besides the practical importance, the problem has also been of great interest in the online algorithms literature (see \citep{Meh13}); online bipartite matching is an actively studied area in algorithm design, e.g., ride-sharing platforms~\citep{Dickerson17,AshlagiBDJSS19}.

We demonstrate the success of our methods for the AdWords problem both for the worst-case and for the robust distributional setting. For the worst-case setting, we provide evidence in two main ways. Firstly, we show that the CR converges empirically to the correct optimal value. To be precise, since it is not tractable to certify the worst-case CR empirically (as that needs evaluation over all inputs of all sizes), we show that the learned algorithm achieves near-optimal competitive ratio over learned as well as known hard instances. Secondly, we show that the learned algorithm shares many salient features of the optimal algorithm: it infers both the \emph{Greedy strategy} and the \emph{Balance strategy}~\citep{KP00}, and seems to find the correct trade-off between them as does the optimal algorithm. 
From the adversary side, the learned input instances successfully defeat several reasonable (but non-optimal) approaches like Greedy and the algorithm optimized on a fixed distribution (\Cref{sec:adwords}). 
For the robust stochastic setting, we show that the learned algorithm smoothly interpolates between the optimal algorithm for the fixed distribution and the worst-case optimal algorithm; by doing so it remains competitive in all regimes --- when the input follows the fixed distribution and when it deviates arbitrarily --- while the optimal distribution-specific algorithm and the worst-case algorithms perform well only in the regime they were designed for. 

\textbf{Generalizability.} While we focus solely on AdWords, we note that our high-level framework can, in principle, be generalized to find algorithms for any optimization problem (online or offline). Not surprisingly, evaluating on a new problem requires designing a problem-specific suite of representations and tests, and it takes considerable ingenuity and engineering effort to implement our framework on a different problem. However, our framework breaks the fundamental limit of previous approaches whose effectiveness crucially relies on strong training sets tuned with domain knowledge which is likely to be available only on well-understood problems. 

Nevertheless, in Appendix~\ref{sec:appendix-skirental}, we take an initial step in this direction, and show that the framework can be used to derive algorithms for classic {\em ski-rental problem\/}~\citep{Karlin86}, matching the CR and the behavior of the optimal randomized algorithm.
Indeed, the larger ambition of our program is to design algorithms for problems for which the literature doesn't have satisfactory solutions; this will need considerably more innovation both to evaluate and to interpret the derived algorithms, and we leave this for future work.

\noindent\textbf{Related work.} The most relevant work is an approach in \cite{kong2018new} that uses reinforcement learning (RL) to find competitive algorithms for online optimization problems including the AdWords problem, online knapsack and the secretary problem. The training there is not from scratch, but based on specific distributions known to be hard for the problem. Specifically, they introduce the notions of universal and high-entropy training set that are tailor-made to facilitate the convergence to worst-case competitive algorithm. In this context, our work takes a major step by eliminating the need for this problem specific knowledge. Further, as we explain in Section~\ref{sec:game-theory}, after viewing the problem of finding worst-case or robust algorithms with the lens of Game Theory, one can see that the previous approach --- significant as it was --- had some gaps. \wadi{This is slightly expanded.} Essentially, the algorithm will suffer from over-fitting to the fixed training set, and thus although it performs competitively on the hand-crafted hard instances, it is not robust to different (and even much easier) examples. Our result overcomes this by addressing the harder problem of learning tabula rasa.

Besides this paper, there are several other recent results on the application of deep learning to solve combinatorial and other algorithmic problems, including \citep{deudon2018learning, kool2018attention, bello2016neural, Dai17, boutilier2016budget, Duetting0NPR19, vinyals2015pointer, graves2014neural, kaiser2015neural, gui2020review}; we defer the discussion of this related work to \Cref{se:full-related-work} but mention that besides \citep{kong2018new}, the other work is different as it does not tackle the problem of finding worst-case or adversarially robust algorithms.

\section{The \YL~framework}\label{sec:yaogan}


We next describe how we implement and stabilize adversarial training for finding max-min strategies through a training framework that we call \emph{\YL} due to the inspiration from Yao's Lemma (Appendix~\ref{sec:game-theory}). 
However, a straightforward implementation of the classic max-min strategy algorithm of \citep{FS96} will be infeasible since finding the best response is often a hard non-convex global optimization problem. Indeed, adversarial responses for the problem require complicated structures to be inferred. One needs to develop a sound strategy to finding good adversarial responses and a mitigation strategy when the search is unsuccessful. 



\begin{figure}
  \centering
  \resizebox{0.55\textwidth}{!}{
  \begin{tikzpicture}
    \node[draw, text width=1.8cm, text centered] at (0,0) (ALG) {Algorithm Network};
    \node[draw, text width=1.8cm, text centered] at (-4,-4) (EXP) {Experience Array};
    \node[draw, text width=1.8cm, text centered] at (4,-4) (ADV) {Adversary network};

    \path (ALG) -- node[draw] (E1) {Evaluation $V(\cdot, \cdot)$} (ADV);
    \draw (ALG) -- (E1);
    \draw[->, -{Latex[width=3mm]}] (E1) --node{Restart + train} (ADV);

    \draw[->, -{Latex[width=3mm]}] (ADV) --node[above]{Append} (EXP);

    \path (EXP) -- node[draw] (E2) {Evaluation $V(\cdot, \cdot)$} (ALG);
    \draw (EXP) --node{Fix best input} (E2);
    \draw[->, -{Latex[width=3mm]}] (E2) --node{Train} (ALG); 
  \end{tikzpicture}}
\caption{The \YL~framework}
\label{fig:yaogan-framework}
\end{figure}

The \YL~framework can be conceptually divided into four components: the algorithm network, the adversary network, the evaluation environment $V(\cdot, \cdot)$ and the experience array (see \Cref{fig:yaogan-framework}). The algorithm network takes the problem-specific online request as input (e.g., arriving bid in the AdWords), and outputs an irrevocable decision (e.g., assignment of the bid in AdWords). Its weights start of at random and are trained until the learned algorithm converges. The adversary network takes random noise as input and outputs the full input (e.g., a complete sequence of bid weights). The weights of the adversary network are frequently reinitialized in order to maximize the probability of finding the worst-case input to a moving algorithm network. Naturally, these two networks require us to model the space of algorithms and inputs with the parameters of the networks.

\textbf{An evaluation environment} evaluates the algorithm network \ALG on an input $i$ and assigns to it a competitive ratio $V(ALG, i)$. We assume in this section that the algorithm wants to maximize the CR and the adversary wants to minimize it. As a reminder, the CR is equal to the ratio of the algorithm's reward to the offline optimum. Specifically, to obtain the algorithm's reward, the evaluation environment feeds the online requests to the algorithm network one-by-one, observes the irrevocable decision, executes the decision and repeats until all requests in the input are exhausted. Furthermore, to obtain the offline optimum, the environment finds the solution with brute force for simple problems or using a linear programming solver (in the case of AdWords).




\textbf{Differentiability issues.} To facilitate backpropagation, we assume that the evaluation measure (namely, competitive ratio) $V(\ALG, i)$ is differentiable --- this applies both to the algorithm's reward (numerator) and to the offline optimum (denominator). This is often a priori not the case and measures need to be taken to appropriately relax the problem. For example, in the AdWords problem the algorithm outputs a discrete bid-to-advertiser assignment value; this discrete assignment is incompatible with differentiability. We circumvent the issue by allowing fractional bid-to-advertiser assignments and show the learned fractional algorithm retains the salient properties of the discrete optimal.
On the other hand, we ensure the differentiability of the offline optimums by computing the optimum (via a brute force search or a linear program solver) for a few perturbed instances and then calculating the (sub)gradient. These issues are problem-specific and we leave them outside of the \YL~framework.

\begin{figure}[t]
\centering
\begin{minipage}{0.6\textwidth}
\begin{algorithm}[H]
    \begin{algorithmic}{\small
        \STATE \textbf{Input:}
        \STATE Differentiable evaluation function $V(\cdot, \cdot)$
        \STATE Parameters $T, \Talg,\Tadv,\Tadd, \Trestart,\nbatch$.
        \STATE {\color{blue}Optional: fixed distribution $D$, parameter $\alpha$}
        \vspace*{2ex}
        \STATE  Initialize algorithm and adversary networks \ALG and \ADV.
        \STATE  Initialize the experience array $E$.
        \STATE  Populate $E$ with $\nbatch$ random inputs.
        \STATE \textbf{For $t=1,\ldots,T$}
            \STATE \textbf{~~~~~~For $u=1,\ldots,\Talg$}
                \STATE ~~~~~~~~~~~~{\color{blue}With probability $\alpha$:
                    \STATE ~~~~~~~~~~~~~~~$I \gets \nbatch$ random instances from $D$.
                    \STATE ~~~~~~~~~~~~~~~Update $\ALG$ via $\nabla_{\ALG} \frac{1}{\nbatch}\sum_{i\in I}V(\ALG, i)$.
                \STATE ~~~~~~~~~~~~With probability $1-\alpha$:}
                    \STATE ~~~~~~~~~~~~~~~$I \gets \nbatch$ random instances from $E$.
                    \STATE ~~~~~~~~~~~~~~~$i^* \gets \arg \min_{i \in I} V(\ALG, i)$
                    \STATE ~~~~~~~~~~~~~~~Update $\ALG$ via $\nabla_{\ALG} V(\ALG, i^*)$.
            \STATE \textbf{~~~~~~For $u=1,\ldots,\Tadv$}
                \STATE{~~~~~~~~~~~~Sample $\nbatch$ random Gaussian tensors $z$}.
                \STATE{~~~~~~~~~~~~Update $\ADV$ via $-\nabla_{\ADV} V(\ALG, \ADV(z))$}.
            \STATE \textbf{~~~~~~If} $t \bmod \Tadd = 0$
                \STATE ~~~~~~~~~~~~Generate $\nbatch$ inputs using \ADV.
                \STATE ~~~~~~~~~~~~Sample $\nbatch$ inputs from $E$.
                \STATE ~~~~~~~~~~~~$I \gets$ the above $2 n_{batch}$ inputs
                \STATE ~~~~~~~~~~~~$i^* \gets \arg\min_{i \in I} V(\ALG, i)$
                \STATE ~~~~~~~~~~~~Append $i^*$ to $E$.
            \STATE \textbf{~~~~~~If} $t \bmod \Trestart = 0$
                \STATE ~~~~~~~~~~~~Reinitialize \ADV to produce random inputs.
    }\end{algorithmic}
    \caption{Generic \YL~Training Paradigm.}
    \label{alg:yaogan}
  \end{algorithm}
\end{minipage}  
\end{figure}

\textbf{An experience array} contains the historical best responses generated by the adversary network. The adversary network goes through cycles of weight re-initialization, training until convergence and storing its final result into the experience array. Moreover, the algorithm network is trained via the experience array and not via the adversary network directly. At each training step of the algorithm, we take some number of samples from the experience array, evaluate the algorithm's performances by computing the CR, find the worst input (smallest CR) among them for the current algorithm, and train the algorithm against that input.


The experience array stabilizes the training and resolves a multitude of issues. The most apparent one is that training the adversary network to convergence takes significantly more time than one step of the algorithm training. Experience arrays amortize the time spent during adversarial training with an equal amount of algorithm network training. Furthermore, an empirical issue that hinders the training is that the adversary network can often fail to find the best response possible. This is not at all unexpected since the question can be intractable in general. Experience arrays allow the adversary networks to use multiple tries in finding the best counterexample. This counterexample might then be used multiple times since one often needs to train on it a few times in succession or revisit it during later training. 

One explanation on why such training is effective is that hard distributions often have small support (a small number of different inputs). This is the case in \cite{kong2018new} where the handcrafted adversarial distributions often comprise a small number of different inputs (although their structure might be complex). Furthermore, even if all sets are of large size, one might still find a satisfactory small-set approximation that converges to the right set.

\textbf{Extension to the robust stochastic setting.} The training algorithm takes an optional distribution $D$ of input instances and a parameter $\alpha \in [0,1]$; in each training iteration for the algorithm network, with probability $\alpha$ we train using samples drawn from $D$, and otherwise we train using samples drawn from the experience array. Moreover, in iterations where we train with samples from $D$, we use the average performance of the algorithm on these instances as the reward function. This extension allows us to learn algorithms in the robust stochastic setting; $\alpha$ controls the trade-off between the distributional and adversarial shock regimes.

\textbf{Pseudocode.} The pseudocode of \YL~is given in \Cref{alg:yaogan}. For the worst-case, \emph{tabula rasa}, training we skip the {\color{blue}optional segments marked in blue}. Robust stochastic setting includes the blue segments (e.g., provides $D$, $\alpha$). We make a final note that our adversary network takes $\nnoise$-dimensional Gaussian noise as input and generates a problem instance. The random noise was typically attenuated during training, but we empirically found that it helped with input space exploration. We use hyperparameters $\nbatch = 100$, $\nnoise = 100$, $\Talg=\Tadv=4$, $\Tadd=100$, and $\Trestart = 100$. The Adam~\cite{kingma2014adam} optimizer is used with paper-suggested parameters. All our experiments were done using TensorFlow 2.0 on a single desktop PC.

\section{AdWords}
\label{sec:adwords}
In this section, we discuss the application of our \YL~ framework to the AdWords problem. 

\textbf{Evaluation Environment.}
Given an instance of AdWords with $n$ advertisers and $m$ ad slots, we slice the input and feed it to the algorithm one ad slot at a time, while keeping track of the allocations made by the algorithm on ad slots already revealed as well as the remaining budgets of advertisers. That is, we make sure the algorithm cannot peek into the future and the decisions are irrevocable. The evaluation environment also computes the total revenue achieved by the algorithm network once the full instance is processed and uses a linear program solver to compute the offline optimum. We use the algorithm revenue and the offline optimum to compute the competitive ratio of the algorithm on the particular instance.

\textbf{Algorithm network.} 
As the evaluation environment feeds the input to the algorithm one ad slot at a time, our algorithm network makes allocation decision for a single ad slot given the current states of the advertisers. Conceptually the algorithm network takes as input $n$ triples (one for each advertiser) that represent the bid $v_i^j$ to the incoming ad slot $j$, the fractional remaining budget $r_i^j / B_i$, and the total budget $B_i$.  It computes, for each advertiser, the probability that the incoming ad slot will be allocated to that advertiser. We note that using the fractional remaining budget (instead of using the absolute value) is for scale-invariance and without loss of generality since it is well understood in theory that the problem retains all its difficulty in the case where all budgets are $1$.

To accomplish this, the algorithm network comprises a single-agent neural network that computes a score for a generic advertiser, and a soft-max layer that converts the scores into probability values.
The single-agent network takes a $6$-dimensional input vector
consisting of the three quantities $v_i^j$, $r_i^j/B_i$, and $B_i$, and the respective sums of these three quantities over all advertisers.
An important consequence of this architecture is that the structure and number of trainable weights is independent of the input size, and the resulting algorithm is \emph{uniform}, that is, it works for arbitrary-sized inputs even though we train it on instances of fixed size.  Additionally, it guarantees that the learned algorithm has the desirable property of {\em permutation equivariant\/}, i.e. permute the advertisers in the input would permute the output probabilities the same way (See \Cref{sec:PE-networks} for details).

During training, we interpret the probability vector output of the network as a fractional allocation of the ad slot. This corresponds to treating the problem as a \emph{fractional} AdWords problem, and ensures differentiability (see Sec~\ref{sec:yaogan}); it is also without loss of generality: the fractional version of AdWords is known to be as hard as the integral version, its solution incorporates the salient features of the integral version. During testing, we interpret the output as actual probabilities and sample one advertiser to allocate to, thus evaluating the original integral version.

\textbf{Adversary network.} The output of the adversary network is an instance of the AdWords problem, which consists of an $m\times n$ matrix $A$. The $j,i$-th entry of $A$ is a real value in $[0,1]$ representing the bid of advertiser $i$ for the $j$-th arriving ad slot. In most of our experiments we fix the budget of each advertiser to be $m/n$, but we also have experiments (in robust stochastic training) where the adversary network additionally outputs an $n$-dimensional vector $B$ of individual budgets. The adversary network takes as input an $m$-dimensional random Gaussian vector to provide randomness to the search for the difficult instances.

Unlike the algorithm network, which is designed to work on arbitrary $m,n$, we do not need the adversary network to scale: in our training, we fix $m=25,n=5$ for the adversary network. 
Since the role of the adversary network is to generate instances to train the algorithm network, we note that, in general, it is desirable for it to generate fixed small-size instances (that capture the essential difficulty of the problem) so that training time/resources may be minimized. 
One may also want to use the adversary-generated instances to gain human-comprehensible intuition about the weakness of an algorithm, and again in this case small but not tiny instances would best serve the purpose.

\subsection{Evaluation}
\begin{table*}[h]
\caption{Revenue of algorithms on triangular and thick-z distribution with $n = 5$ advertisers (higher is better). The budget for each advertiser is the number of ads $m$ divided by $n$, so the offline optimal always equals to $m$ (note: one can easily calculate CR by dividing the result by $m$). We present the average revenue ($\pm$ standard deviation) on $100$ samples from each distribution.}

\label{table:5byx}
\centering
\begin{tabular}{ccccccc}
\toprule
Distr. & Ads ($m$)
& \YL~ & MSVV & Greedy & Fixed-training\\
\midrule 
 & $25$
 & $17.15(\pm 0.09)$ & $17.16 (\pm 0.36)$ & $17.12 (\pm 1.47)$ &$17.14 (\pm 0.45)$ \\
Triangular & $50$ 
& $34.29(\pm 0.14)$ & $34.29 (\pm 0.45)$ & $33.71(\pm 1.94)$ &$34.36(\pm 0.85)$ \\
 & $100$ 
 & $68.64(\pm 0.20)$ & $68.65 (\pm 0.47)$ & $69.44(\pm 2.71)$ &$65.98(\pm 5.71)$ \\
\midrule 
 & $25$ 
 & $17.76(\pm 0.09)$ & $18.01 (\pm 0.59)$ & $15.9 (\pm 1.12)$ &$17.32 (\pm 0.43)$ \\
Thick-z & $50$ 
& $35.31(\pm 0.13)$ & $35.89 (\pm 0.46)$ & $31.59(\pm 1.82)$ &$34.67(\pm 0.78)$ \\
 & $100$ 
 & $70.21(\pm 0.22)$ & $71.83 (\pm 0.37)$ & $61.44(\pm 2.07)$ &$69.47(\pm 6.28)$ \\
\bottomrule
\end{tabular}
\end{table*}

\noindent

We present empirical evaluations to demonstrate the effectiveness of our framework. 
We begin by observing that there is no known tractable way of certifying the competitive ratio of a given algorithm in general --- a brute force approach is infeasible even for $25 \times 5$ instances with $\{0, 1\}$ bids. Furthermore, formalizing the certification task yields a highly non-convex optimization problem, which we have no reason to believe is tractable.
Instead, we evaluate the learned networks it in several ways --- measuring the algorithm's performance on known hard benchmark inputs (Takeaway 1); tracing how the adversary network guides the algorithm's evolution towards an empirical CR that matches the optimal CR known in theory (Takeaway 2); testing whether the algorithm network has learned the ingredients of the optimal solution, namely, Greedy and Balance strategies (Takeaway 3).  In addition, we evaluate whether the algorithm scales to input sizes beyond what it was trained for (Takeaway 4), and we present a number of results in the robust stochastic setting (Takeaway 5).

\textbf{Takeaway 1: \YL~ finds an algorithm that is competitive with the optimal worst-case algorithm (MSVV) and strictly dominates the natural Greedy strategy on hard benchmark distributions; it also matches the algorithm explicitly trained on the benchmark distributions (``Fixed-training''), without requiring any knowledge of the hard distributions.} This is showcased in Table~\ref{table:5byx}.

{\bf Greedy} is the simple strategy that assigns an ad slot to the advertiser with the highest bid (truncated by the advertiser's remaining budget), ties broken randomly. While Greedy is natural and effective, it is sub-optimal --- this can be seen on the thick-z distribution where it gets significantly weaker performance than the other strategies, including \YL~.

\noindent
{\bf MSVV} is the optimal worst-case online algorithm for AdWords~\cite{MSVV07}. %
For the $j$-th arriving ad, the algorithm scales advertiser $i$'s bid $v^j_i$ to be $q^j_i := v^j_i (1-\exp(- s^j_i))$ where $s^j_i := r^j_i / B_i$ is the fraction of advertiser $i$'s budget remaining when ad $j$ arrives. The algorithm then runs Greedy on the scaled bids. In the unweighted case (i.e. bids are $\{0,1\}$), the strategy is known as {\em Balance} as it assigns ads to the advertiser with most remaining balance among those bidding $1$. The algorithm always guarantees at least a $(1-1/e) \approx 0.632$ fraction of the offline optimal revenue (with the small-bids assumption that we also make), which is asymptotically optimal for large $n$ and $m$; smaller instances can have better CRs. We see that the algorithm found by \YL~ is competitive with MSVV on the two tested input distributions.


{\bf Fixed-training.} We recreate the general outline of the state-of-the-art in algorithmic learning for AdWords as described in~\cite{kong2018new}. Namely, we fix the input distribution to be a mixture of of the triangular and thick-z inputs (discussed below) and train the algorithmic network for it.  Unsurprisingly, the resulting algorithm does well on both distributions; note that \YL~ matches its performance, without being trained on these benchmark distributions (nor any human-picked distributions).

\noindent
{\bf Input Distributions.}
We use the distribution constructed from the triangular graph and the thick-z graph (see \Cref{sec:app-adwords-inputs} for details) as the benchmark (same as~\cite{kong2018new}). The triangular graph is canonical because a distribution generated by randomly permuting the columns of the triangular graph is the minmax input distribution for the AdWords problem --- i.e., no online algorithm can achieve a CR greater than $1 - 1/e$.
The thick-z graph is also canonical --- column permutations of the thick-z graph are particularly bad for Greedy, leading to a CR of $0.5$ asymptotically (it is employed by~\cite{kong2018new} in their training to prevent the network to just learn Greedy).


\begin{table}[h]
\caption{The revenue of \YL~ that was trained on $25 \times 5$ instances evaluated against larger $m\times n$ instances. Uniform budget is set to $m/n$ so offline optimum equals $m$. Average revenue ($\pm$ std) over $100$ samples shown.}
\label{table:largern}
\centering
\begin{tabular}{ccccc}
\toprule
Distr. &  $m\times n$  & \YL~ & MSVV & Greedy \\
\midrule 
Triangular & $100\times 10$ & $66.11(\pm1.18)$ & $66.27(\pm 0.97) $ & $65.47(\pm 2.92)$  \\
 & $400\times 20$ & $259.13(\pm 2.51)$ & $259.26(\pm 0.97)$ & $259.40(\pm 5.64)$ \\
\midrule 
Thick-z & $100\times 10$ & $68.48(\pm 1.37)$ & $70.02(\pm 0.63) $ & $58.54(\pm 2.49)$ \\
 & $400\times 20$ & $260.48(\pm 2.75)$ & $277.02(\pm 1.02)$ & $221.21(\pm 3.80)$ \\
\bottomrule
\end{tabular}
\end{table}

\noindent

\begin{figure}[h]
  \begin{center}  
    \includegraphics[width=0.7\textwidth,height=3.2cm]{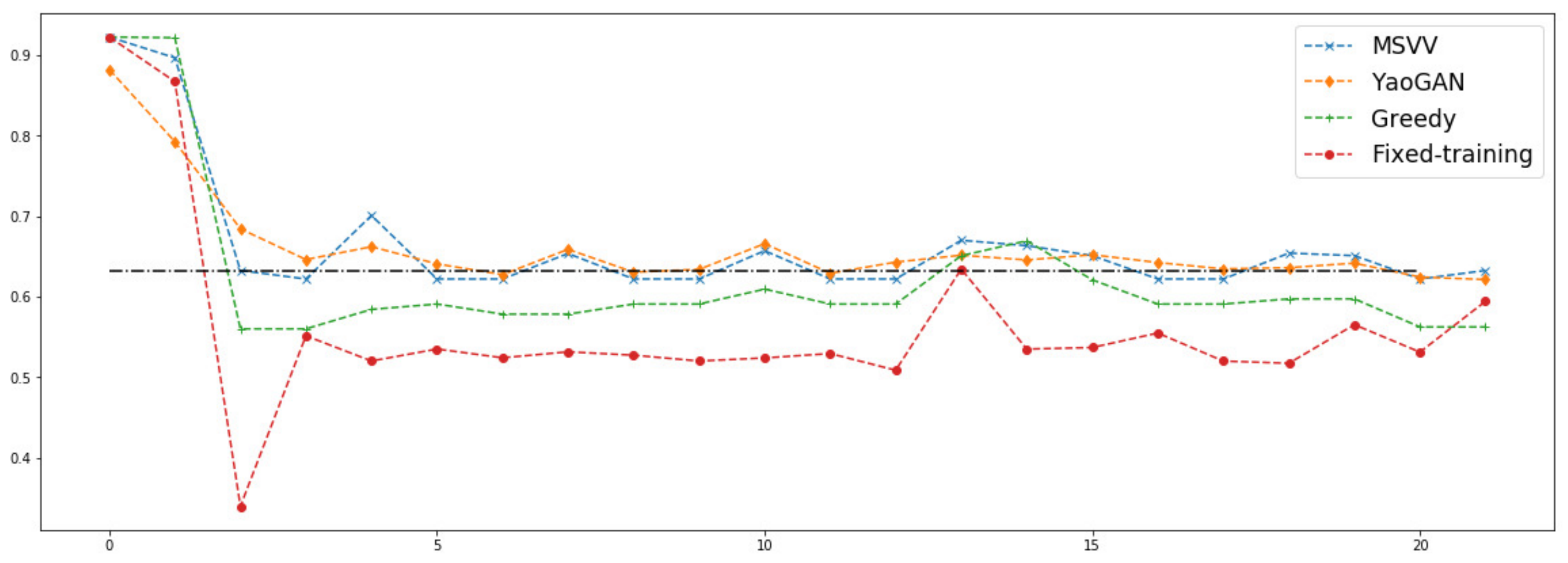}
    \caption{We plot the CRs of various algorithms on the input instances in the experience array generated during \YL~ $m \times n = 25 \times 5$ training. Horizontal axis is the index of the instance in the experience array (if the adversary network works as intended, later instances should be ``harder''). The vertical axis is the CR of algorithms evaluated on these instances. As a reference, the black horizontal line is the optimal worst-case competitive ratio (achieved by MSVV) under the small-bid assumption. There are $250$ instances in the experience array. To smooth out noise, we group every $10$ consecutive instances as a batch, and plot the worst (minimum) CR on instances within each batch for the algorithms. }
    \label{fig:adversary}
  \end{center}
\end{figure}

\textbf{Takeaway 2: The estimated CR of the adversarial training converges to the optimal CR. Moreover, the adversary network generates examples that  are hard for both Greedy and Fixed-training.} 
Here we estimate the CR by measuring the performance on the hardest inputs generated by the adversary network. This is showcased in \Cref{fig:adversary}. It can be seen that the empirical CR of \YL~ converges to the textbook-optimal CR.

We next address the question of whether the inputs generated by the adversary network are merely hard for the algorithm network, or are hard instances of the AdWords problem in a broader sense.  The performance of MSVV on these instances, as shown in
Figure~\ref{fig:adversary}, clearly demonstrate that they force the worst-case performance of MSVV as well.
It also clearly shows \YL~ generates instances of increasing difficulty as the training evolves.
Similarly, 
the plot shows that for Greedy, \YL~-generated instances are more difficult than the triangular graph. Also note the earlier instances are very easy as they are basically random $[0,1]$ matrices. MSVV, Greedy and \YL~ are all able to exploit this and achieve very good competitive ratio. On the contrary, Fixed-training does even worse on these easy instances (comparing to the more difficult ones later), which can be explained since these easy instances look very different from the expert crafted instances used to train Fixed-training. Finally, we demonstrate the effectiveness of the adversary network by {\bf training the adversary against a fixed algorithm}: we effectively find hard examples for both Greedy and MSVV with CR close their theoretical worst-case CR (see \Cref{sec:adv-training-fixed}).



\begin{table*}[h]
\caption{The average CR achieved with robust stochastic training via \YL~ by using the Powerlaw distribution. For training, we used the Powerlaw$(5)$ distribution and trained multiple algorithms by varying the $\alpha$. We refer to these algorithms as pwl-100, pwl-95, pwl-90 where the number indicates the value of $\alpha$ in percentage. E.g., pwl-100 is trained exclusively with Powerlaw instances, and pwl-95 is to use instances from Powerlaw with probability $0.95$. We refer to the algorithm trained exclusively with the adversary network as tabula-rasa.}
\label{table:fulltable}
\centering
\begin{tabular}{ccccccc}
\toprule
Distr. &  MSVV & Greedy & pwl-100 & pwl-95 & pwl-90 & tabula-rasa \\
\midrule 
Thick-z$(5)$ & $0.714$ & $0.636$ & $0.622$ & $0.687$ & $0.708$ & $0.714$\\
Powerlaw$(5)$ & $0.939$  &  $0.993$  & $0.989$ & $0.985$ & $0.984$ & $0.878$\\
Triangular-g$(5)$ & $0.861$ & $1.000$ & $0.997$ & $0.990$ & $0.984$ & $0.814$\\
\midrule
Thick-z$(10)$ & $0.700$ & $0.589$ & $0.568$ & $0.667$ &  $0.706$ & $0.685$\\
Powerlaw$(10)$ & $0.956$ & $0.986$ & $0.977$ &  $0.977$ & $0.970$ &  $0.908$ \\
Triangular-g$(10)$ & $0.847$ & $0.998$ & $0.978$ & $0.981$ & $0.967$ &  $0.844$ \\
\bottomrule
\end{tabular}
\end{table*}

\textbf{Takeaway 3: The algorithm learned by \YL~ discovers the salient features of MSVV.} In Appendix~\ref{sec:app-adwords}, we further test the learned network using inputs crafted to create various critical scenarios. The results demonstrate that our algorithm follows very accurately the optimal strategy in these controlled settings. We provide a sample preview of these results in Fig~\ref{fig:yaogan-behavior-sample}.

\textbf{Takeaway 4: the \YL~ scales to larger instances than it was trained on.} This is showcased in Table~\ref{table:largern}: We perform the same \YL~ training as described before (on instances of dimension $m \times n = 25 \times 5$), but then measure its performance on instances of larger size.
Notably, \YL~ still maintains performance somewhat similar to the optimal worst-case MSVV and still clearly outperforms Greedy (on the thick-z distribution). This suggests that this approach of training on small examples and scaling is effective and provides multiple benefits: small examples can be easily interpreted and intuition can be garnered from them and the training is significantly faster.

\textbf{Takeaway 5: \YL~ enables robust stochastic training to find algorithms that are both highly attuned to the distribution of interest and are robust to adversarial shocks.}
We use the extension of the \YL~ framework to perform robust stochastic training. We specify a distribution of interest $D = \text{Powerlaw}$ (see below) and train multiple algorithms by varying $\alpha$. We measure the performance against the following three distributions:

\begin{figure}[hpbt]
\centering
\begin{subfigure}{.45\textwidth}
  \centering
  \includegraphics[width=.75\linewidth]{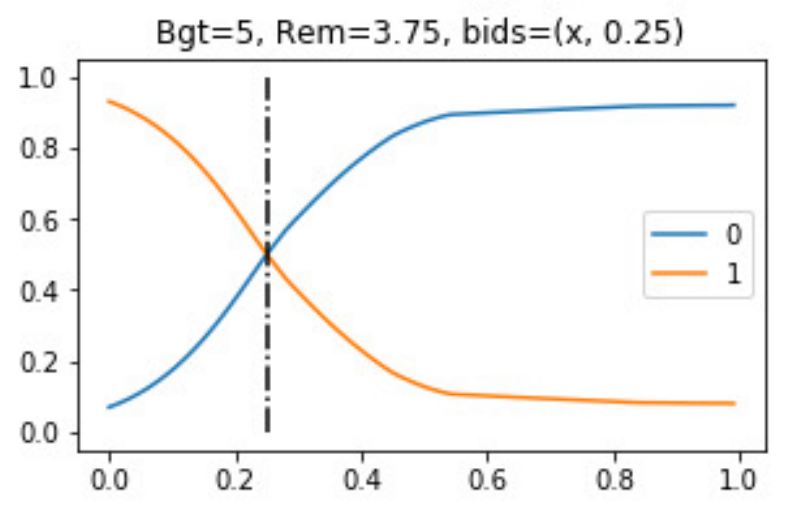}
  \caption{\YL~ correctly exhibits Greedy behavior when advertisers have the same (remaining) budget: the MSVV response is to take the highest bid.}
\end{subfigure}%
\begin{subfigure}{0.05\textwidth}
  \centering
  \phantom{.}
\end{subfigure}
\begin{subfigure}{.45\textwidth}
  \centering
  \includegraphics[width=.75\linewidth]{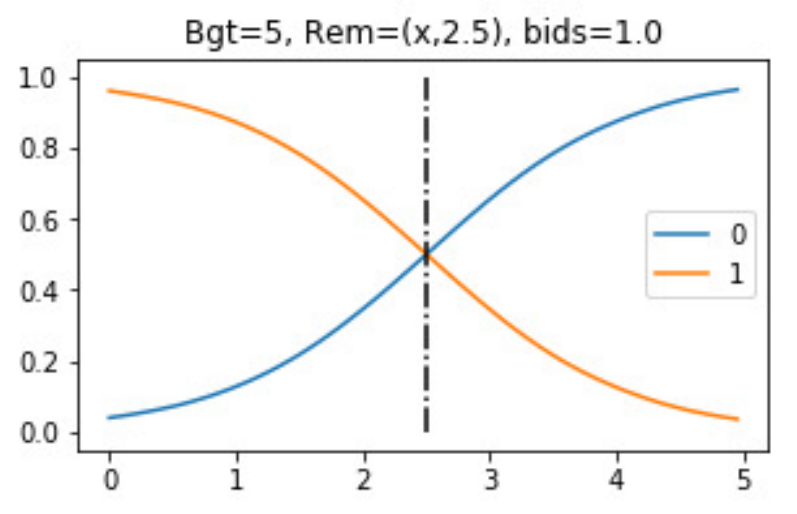}
  \caption{\YL~ correctly exhibits Balance behavior on equal bids: the MSVV response is to assign the ad to the highest remaining budget advertiser.}
\end{subfigure}
\caption{We dissect \YL~-trained algorithm and compare its behavior with MSVV in two regimes: the Greedy and Balance regime. In 3(a) we set up advertisers $0$ and $1$ with a fixed common remaining budget, vary $0$'s bid ($x$-axis) while fixing $1$'s bid (the black dotted line), and plot our algorithm's output allocation probabilities ($y$-axis). In 3(b) we fix the bids and vary the remaining budget for advertiser $0$. In both scenarios the (optimal) MSVV algorithm has a sharp $0 \to 1$ transition at the dotted line for advertiser $0$. In fact, an optimal algorithm for the fractional AdWords problem has allocation curves very close to the trained ones.}
\label{fig:yaogan-behavior-sample}
\end{figure}


    
    \emph{Thick-z}$(n)$: the $n^2\times n$ sized (i.e. $m=n^2$) thick-z graph, and all advertisers have budget $n$.
    
    \emph{Powerlaw}$(n)$: this distribution is inspired by high-level characteristics of real-world instances, generated by a preferential attachment process on $n^2\times n$ sized graphs.
    Further, we set the budgets in a way that Greedy obtains the offline optimum, which also makes the MSVV algorithm overly conservative in its hedging (details are deferred to \Cref{sec:app-adwords-inputs}).
    
    \emph{Triangular-g}$(n)$: the $n^2\times n$ sized triangular graph. Each non-zero bid is drawn i.i.d from the uniform distribution over $[0.5,1]$, and the bidders' budgets are set so the simple greedy strategy would achieve near perfect competitive ratio similar to the case of Powerlaw.
    


The robust stochastic training leads to algorithms (e.g. pwl-90) that are (near) optimal on both the distribution they were trained on (i.e. Powerlaw) and Triangular-g which, by the design of budgets, shares the same optimal Greedy strategy with Powerlaw. This indicates that robust stochastic training is effective in exploiting the correlations of the distribution it was trained on. Furthermore, as we decrease $\alpha$ (i.e., incorporate more adversarial samples), we see that the learned algorithm interpolates between the performance of tabula-rasa (trained exclusively with the adversary network) and $\alpha = 1$ (trained exclusively with Powerlaw). This indicates that the algorithm becomes more robust to adversarial shocks, as showcased in Thick-z of \Cref{table:fulltable} where including more adversarial training is effective at finding more robust algorithms that can avoid the common pitfalls of Greedy on these canonical adversarially constructed bad instances. 



\section{Conclusion}
We explore the idea of using adversarial training to learn algorithms that are robust to worst-case inputs. The distinct challenge is to design a paradigm that can work from first principles without any problem-specific expert knowledge, and we demonstrate the effectiveness of our framework on the AdWords problem. The ultimate goal of our investigation is to employ the power of machine learning to gain critical insights and to aid the design of new algorithms beating state-of-the-art on interesting open problems in previously unexplored ways. While our work is very preliminary, we make a solid step towards this long term goal by removing the fundamental limit of prior approaches requiring existing problem-specific expertise. We hope our research can inspire more work from both the ML side and algorithm side along this direction.

\textbf{Acknowledgement.} The authors would like to thank Manzil Zaheer for many helpful discussion that shaped the early version of this paper. We would also like to thank the anonymous reviewers whose suggestions greatly improved both the presentation and content.





\appendix
\newpage 

\section{Related Work}\label{se:full-related-work}

The most relevant work is an approach in \citep{kong2018new} that uses reinforcement learning (RL) to find competitive algorithms for online optimization problems including the AdWords problem, online knapsack, and the secretary problem. The training there is not from scratch, but based on specific distributions known to be hard for the problem. Specifically, they introduce the notions of universal and high-entropy training set that are tailor-made to facilitate the convergence to worst-case competitive algorithm. In this context, our work takes a major step by eliminating the need for this problem specific knowledge. Further, as we explain in Section~\ref{sec:game-theory}, after viewing the problem of finding worst-case or robust algorithms with the lens of Game Theory, one can see that the previous approach --- significant as it was --- had some gaps, which are overcome by addressing the harder problem of learning tabula rasa.

Among other work on using ML to train algorithms (which are more loosely related to our work), \cite{deudon2018learning} and \cite{kool2018attention} use RL to solve the traveling salesman problem (TSP) in the two-dimensional plane. Both uniformly sample the input points from the unit square and teach the network to optimize the expected performance on this input distribution without having access to exact solutions. Nevertheless, the learned algorithms from both approaches seem competitive with state-of-the-art methods, at least for modest input sizes. \cite{vinyals2015pointer} introduces the Pointer Network, a sequence-to-sequence model that successfully manages to approximately find convex hulls, Delaunay triangulations and near-optimal planar TSPs. They assume access to an exact solution and use supervised learning to train the model, while their input is uniformly sampled from the unit square. \cite{graves2014neural} introduce the Neural Turing machines, a differentiable and computationally universal Turing machine model, and use supervised learning to teach them basic algorithms such as copying or sorting. \cite{kaiser2015neural} introduce the parallel and computationally universal Neural GPU and use supervised learning to teach it addition and multiplication that seemingly generalizes to arbitrary lengths. 

We point out some other related results without going into details: the work of~\cite{bello2016neural} studies applying RL on combinatorial problems including TSP and Knapsack; the work of~\cite{Dai17} exploits embedding and RL to design heuristic algorithms for classic graph problems on specific distributions;~\cite{boutilier2016budget} uses RL to solve Budget Allocation problem;~\cite{Duetting0NPR19} applies deep learning techniques to designing incentive compatible auction mechanisms. 

To the best of our knowledge, previous results either learn algorithms designed to work on specific input distributions of interest, or utilize supervised learning to mimic a known method, or (in the case of~\cite{kong2018new}) rely on domain expertise to provide hard input instances to train algorithms that are competitive in worst-case. For the specific problems we consider, readers interested in ski rental and AdWords can consult~\cite{Buchbinder:Naor:survey} and \cite{Meh13} for a more thorough coverage.

\section{Online Algorithms}
\label{app:theory}
In this section, we provide more background on the essential notions of online optimization for a broader audience.

Online algorithm is a very different computational model from the classical (offline) algorithm design, and arises broadly in practice in situations where one has to make decisions in real time without knowing the future. 

In the online context, the input is revealed sequentially instead of all at once, and the algorithm needs to make irrevocable decisions during the process using the input revealed so far similar to the Online Learning problem. The difficulty comes from making irrevocable decisions without knowing future input, while the current decision will have implications in the future. Take AdWords as an example: the search engine can sell an ad slot (i.e., user impression) each time some user searches for some query. Advertisers specify their budgets (i.e., total amount of money they are willing to spend on all ads) at the beginning of the episode and then place bids each time a new ad slot arrives. The search engine has to decide in real time how to allocate the ad slot to advertisers given their bids and budgets. The goal for the search engine is to maximize its revenue over the entire input sequence. Once we display the ad from an advertiser, there is no way to change the decision later, and once the budget of an advertiser is depleted, the search engine cannot sell ads to the advertiser any more. 

\smallskip

\noindent{\bf Inherent difficulties in online optimization.} Consider the instance with two bidders $A,B$ both having budget $1$, and two ad slots arriving sequentially. Suppose $A$ and $B$ both bid $1$ for the first ad. To whom do we allocate the ad? With out loss of generality, suppose we give the first ad to $A$, and collect payment of $1$ from $A$. Now suppose the second ad slot comes, and only $A$ bids $1$ for it (and $B$ bids $0$). We cannot collect any more revenue from $A$ as $A$’s budget is already spent on the first ad, and cannot collect any revenue from $B$ since $B$ values the second ad at $0$. In this case the total revenue we get is $1$. However, hypothetically, if we knew the entire input sequence before making any decision, we could have assigned the ads optimally (the first one to $B$ and the second to $A$), getting a total revenue of $2$. This revenue is called the \emph{offline optimal}, i.e., how well we could have done in retrospect. Note that in the AdWords problem as typically considered, we assume that each bid is small compared to the budget; the above example does not follow this assumption but is easier to understand (it is an instance of the Online Bipartite Matching problem). However, even with the small-bids assumption, we have essentially the same issue if we consider $A$ and $B$ to have budgets of $100$ each, and two groups of a $100$ ads arriving similarly. Further, the differences in the bid values create even more possibilities of incorrect decisions.

\smallskip

\noindent{\bf Competitive ratio (CR).} 
The primary measure of the performance of online algorithms is the \emph{competitive ratio} (CR). For a fixed input instance this is the ratio of the (expected) objective value the online algorithm achieves for that instance to the \emph{offline optimal value} (the best possible objective value for the instance, knowing the entire sequence upfront). In the above example, we get a competitive ratio of $1/2$ on that particular instance for that algorithm. The competitive ratio of the algorithm is defined as its worst competitive ratio over all possible inputs. For the Adwords problem it is known that there is a deterministic algorithm (from~\citep{MSVV07}) achieving a CR of $1-1/e$ with the small-bids assumption.

\smallskip

\noindent{\bf Adversarial model.} There are various models specifying what assumptions we have on inputs (i.e., on what inputs we want the algorithm to achieve good CR), e.g. i.i.d, random order, or adversarial. In this paper we assume the, most general, \emph{adversarial model}. In it, we want the algorithm to achieve good CR against all possible inputs, and we refer to the CR of an algorithm as the worst CR the algorithm gets on any input. This model achieves the highest level of robustness since it does not put any assumptions on the input; making it a meaningful model to consider in mean real-world settings. For example, in online advertising, the ad slots supply (i.e. user searches) can change drastically when some events go viral. Moreover, advertisers spend a great amount of resources to reverse-engineer the search engine's algorithm and come up with sophisticated bidding strategies in response. So it is of great interest to design algorithm that performs well no matter how the inputs look like.

\noindent{\bf Robust stochastic model.} In a stochastic model, the inputs of interest come from a fixed distribution (which can be known in advance, or learned over time). It is known (see, e.g.,~\citep{Meh13}) that one can develop algorithms with much better performance (CRs), as expected. In this paper, we also study the \emph{Robust Stochastic Model}.
This model is motivated by the example mentioned above in which an algorithm designer may have very good estimates of the incoming data (lending itself to a stochastic model), but there can be arbitrary jolts or shocks in the system (e.g., some viral event), which can lead to completely different inputs. In the robust stochastic model, one wishes to design algorithms which perform really well (better than the CR for the adversarial model) when the inputs follow the estimates, but also degrade smoothly and have good guarantees when the inputs deviate adversarially.


\section{The Game theoretic framing}
\label{sec:game-theory}

\subsection{The game-theoretic approach to adversarial training}
We start with a short discussion on how adversarial training can be seen through the lens of game theory. 

One can frame the task of finding an optimal algorithm for a given problem in an abstract game-theoretic setting: Consider a (possibly infinite) two-player zero-sum game matrix $V$, in which
each row represents a deterministic algorithm $\mc{A}$, and each column represents an input instance $\mc{I}$. The entry $V(\mc{A},\mc{I})$ is the payoff (CR) to Row if it plays $\mc{A}$, and Column player plays $\mc{I}$. 
The worst-case optimal (randomized) algorithm corresponds to Row's max-min strategy. By (von Neumann) strong duality,
Column's min-max strategy is therefore the worst (most difficult) input distribution for this problem (this application of duality is called Yao's Lemma~\citep{Yao77} in Theoretical CS). 

\textbf{Yao's Lemma.} Online algorithms are often randomized since hedging between different input continuations can be elegantly achieved via randomization. 
A randomized algorithm can be viewed as a distribution $\vp$ over rows of $V$ (i.e., deterministic algorithms); an input distribution $\vq$ can be viewed as a distribution over columns of $V$ (i.e., input instances). The optimal (potentially randomized) algorithm $\vp^*$ is one whose expected payoff against the worst-case input is minimized, i.e., $\vp^* = \arg\min_{\vp} \max_{\mc{I}} \vp^T V(:,\mc{I})$. A standard result in game-theory stipulates that once you fix a randomized algorithm, then maximizing over pure instances (columns) is equivalent to maximizing over input distributions. The namesake of our paper, Yao's Lemma~\citep{Yao77} says that there exists a distribution $\vq^*$ over inputs such that the expected payoff of the best (deterministic) algorithm over $\vq^*$ is the same as the value of the best randomized algorithm $\vp^*$ over its worst possible (pure) input. In other words, $\max_{\mc{I}} (\vp^*)^T V(:, \mc{I}) = \min_{\mc{A}} V(\mc{A}, :) \vq^*$.  This is, of course, simply an elegant application of von Neumann's minmax theorem aka strong duality to this specific game. Using classic game theory notations, we also refer to the best randomized algorithm as the row player's minmax strategy, and $\vq^*$ the column player's maxmin strategy. Solving an online optimization problem corresponds to solving the above zero-sum game, specifically finding $\vp^*$.

Prior work in~\citep{kong2018new} used this to train an algorithm network on the min-max input distribution for the Adwords problem. Besides the drawback that this needs problem expertise to know the min-max distribution, this approach also has a fatal flaw that Row's best response  to a min-max strategy of Column need not be its own max-min strategy (e.g., in Rock-Paper-Scissors, any pure strategy is a best response to the max-min strategy of playing the three moves uniformly at random). Hence, the worst input distribution is not sufficient to recover the best algorithm; this is specifically true for the AdWords problem. Indeed, this issue was acknowledged in~\cite{kong2018new}: they combine multiple types of special input distributions (e.g., universal and high-entropy training sets).
To overcome this obstacle, we note that solving the game from scratch necessarily requires solving for both the algorithm and the adversary. We draw inspiration from online learning for game-solving \citep{FS96}, and show how we can view adversarial training in this manner. Adversarial training intuitively seems to be the ``right thing'' to train algorithms with worst-case guarantees, and we show that the unique challenge now is co-training the Adversary as well, which extends and strengthens the existing adversarial training framework.

\subsection{No regret dynamics and convergence to the MinMax}
One way to explain the success of \YL~ is in that it loosely follows no-regret dynamics. There is a rich line of work on no regret dynamics~\citep{FS96,AHK} which lies at the center of various fields including game theory, online learning and optimization. In the context of two-player zero-sum games, this gives a method to approximately solve the value of the game without knowing the optimal column player strategy (i.e., the adversarial input distribution) in advance. We extend the notation to use $V(\vp,\vq)$ to denote $\vp^TV\vq$.
  
More specifically, we can let the algorithm player maintain a randomized algorithm $\vp^t$ over iterations $t=1,\ldots,T$. In each step the adversary plays the best response $\mc{I}^t$ (i.e., the worst-case input against the algorithm at that iteration). The algorithm $\vp^{t+1}$ is then computed from $\vp^t$ and $\mc{I}^t$ using a specific no-regret dynamic such as Hedge~\citep{FS96}. Define
\[
\tilde{\vp} = \argmin_{\vp^t}V(\vp^t,\mc{I}^t),\qquad 
\tilde{\vq} = \frac{1}{T}\sum_t \mc{I}^t .
\]
Classic results give us that the payoff of the pair $\tilde{\vp},\tilde{\vq}$ converges to the optimal minmax value for sufficiently large $T$.

Our \YL~ training loosely follows this approach. It maintains a randomized algorithm $\vp^t$ and searches for the best response $\mc{I}^t$. However, it does not follow the no-regret dynamics of prior algorithms, but rather uses simple gradient descent to optimize the payoff of the algorithm on $\mc{I}^t$. 



\section{Ski Rental}\label{sec:appendix-skirental}
%
%
%

In this section, we describe how our framework can be applied to find a worst-case competitive algorithm for a different online optimization problem, Ski-Rental. We describe the problem, the techniques used, and the results which demonstrate that the framework has found a strategy extremely similar to the theoretically optimal randomized algorithm.
\subsection{Problem definition}\label{sec:def-skirental}
We formally define the ski rental problem and illustrate the notions of deterministic strategies and competitive ratios.

\begin{problem}[Ski rental]
    Suppose we want to go skiing in the winter, which requires a set of skis. Each day we go skiing, we have two options: either rent the skis for $\$1$, or buy the skis for  $\$B$, for some fixed $B$. Naturally, renting the skis allows us to ski only for that one day, while after buying we can use them for an unlimited number of days. The crux is that we do not know in advance the total number of days $k$ we will be able to ski (e.g., we might get injured or have some work come up, etc.). The objective is to minimize the total amount of money spent by strategically buying/renting skis.
\end{problem}

\noindent{\bf Deterministic algorithms.} A deterministic algorithm $\mc{A}$ takes the form of renting for up to some fixed number $x$ of days initially, and if it turns out that we go skiing for at least $x$ days, we buy on day $x$. If the eventual number of days we go skiing is $k$, the cost of this strategy will be $\min(k, B+x-1)$, and the offline optimal cost will be $\min(k,B)$. The corresponding entry in the payoff matrix $V$ is $\frac{\min(k,B+x-1)}{\min(k,B)}$, and the competitive ratio of this deterministic algorithm is $\max_{k\geq 1} \frac{\min(k,B+x-1)}{\min(k,B)}$, i.e., for the worst possible input $k$. 

\noindent{\bf Randomized algorithms and the optimal algorithm.} As opposed to the AdWords problem, one interesting aspect here is that the optimal algorithm is randomized. A randomized algorithm for the ski rental problem is represented by a (marginal) probability distribution $(p_1, p_2, \ldots, p_N)$, with the semantics that for each $x \in \{1, 2, \ldots, N\}$, the value $p_x$ denotes the probability that the we rent skis for the first (up to) $x-1$ days and buy skis on day $x$, if the season extends that far. In the most interesting case $N \to \infty$ we note that the optimal choices are given by $p_i = (\frac{B-1}{B})^{B-i}\frac{c}{B}$ for $i=1,\ldots,B$, where $c=\frac{1}{ 1 - (1 - 1/B)^B}$ so these probabilities sum to $1$. It has a competitive ratio of $\frac{1}{1 - (1 - 1/B)^B}$, which tends to $\frac{e}{e - 1} \approx 1.582$ as $B \to \infty$~\citep{Karlin86}.

\subsection{Learning the optimal algorithm via adversarial training}
In this section, we show how the framework finds a strategy very close to the textbook-optimal randomized algorithm. 

\textbf{Representation.} Our goal here is to learn an algorithm that works for {\em all} values of $B$ and $N \to \infty$. This is out of reach for traditional optimization methods for a number of reasons, the most obvious one being the infinite number of variables one needs to optimize over. As $B$ and $N$ can be (asymptotically) large, we instead work with a continuous space, where every parameter is essentially normalized by $N$. In this setting, time goes over the range $[0,1]$ indicating what fraction of the time horizon has passed. At time $\tau\in [0,1]$, if we are still skiing, the cost of renting from the beginning of time up to this point is captured by $\tau$. The cost of buying is captured by a real number $\beta\in [0,1]$, which is the fraction of the time one needs to go skiing in order to make buying the skis worthwhile. If the season ends at point $\alpha$ of the continuous timeline, the optimal offline cost is $\min(\alpha,\beta)$. The algorithm's cost is $\beta + \tau$, if we choose to buy at point $\tau\in [0,1]$ where $\tau\leq \alpha$, and the cost is $\alpha$ if we don't buy before point $\alpha$. 
The algorithm network takes $\beta$ as input, and outputs a randomized algorithm for the particular $\beta$ as a cumulative distribution function (cdf) of buying probabilities over time.
We access the algorithm by querying the cdf.
To instantiate the algorithm for a particular pair of $B,N$ values, we run it with $\beta=B/N$, and query the returned cdf at $\tau = i/N$ for $i=1,\ldots,N$.

\textbf{Adversarial training.}
For the ski rental problem, we apply a subset of the full \YL~ framework. Specifically, we train a deep neural network to represent the algorithm, but for the adversary, we don't resort to deep neural networks. This decouples the algorithmic from the adversarial training and showcases that best-response training is effective. 

The training goes as follows: fix an algorithm network, find (via a search) the worst-case input in terms of CR to this fixed network, optimize the algorithm network on this input and repeat. The {\bf algorithm network} is a deep neural network that takes $\alpha,\beta\in[0,1]$ as inputs, computes the CDF of a probability distribution $p_\beta$, and outputs the value of the CDF at $\alpha$. As for the architecture of the algorithm neural network, we use $50$ Gaussian kernels $N(x_i,\sigma)$ for $i=1,\ldots,50$ with fixed means (i.e., $x_i$'s equally spaced over $[0,1]$) and standard deviation $\sigma=2/50$, and take a weighted average of these Gaussian kernels as the distributions learned by our neural network. In particular, given any $(\alpha,\beta)$ as input, the neural network uses $\beta$ to derive weights $w_i$'s over the Gaussian kernels (where $\sum_i w_i=1$), and returns the weighted sum of the CDF of these fixed Gaussian kernels at point $\alpha$ as the output of the algorithm. We use a fairly generic neural network structure to learn the kernel weights: the network takes $(\alpha, \beta)$ as inputs and first applies 4 hidden ReLU-activated layers with 256 neurons. The last hidden layer, the one that outputs $w$, is a softmax-activated. Adam optimizer~\citep{kingma2014adam} with the default parameters is used to train the network

On the {\bf adversary} side, in each training iteration, we randomly sample a small number of values for $\beta$, and for each sampled $\beta$ we use an $\epsilon$-net over $[0,1]$ as values of $\alpha$ with $\epsilon$ chosen to be $0.01$. We approximately evaluate the competitive ratios (up to certain precision) of the current algorithm for these $(\alpha,\beta)$ instances, and train the algorithm neural network with the instance where it gets the worst competitive ratio. The approximation is due to not really taking a integral over the learned cdf but rather a step function approximation of it.

\begin{figure}[h]
  \begin{center}  
    \includegraphics[width=\textwidth,height = 6cm]{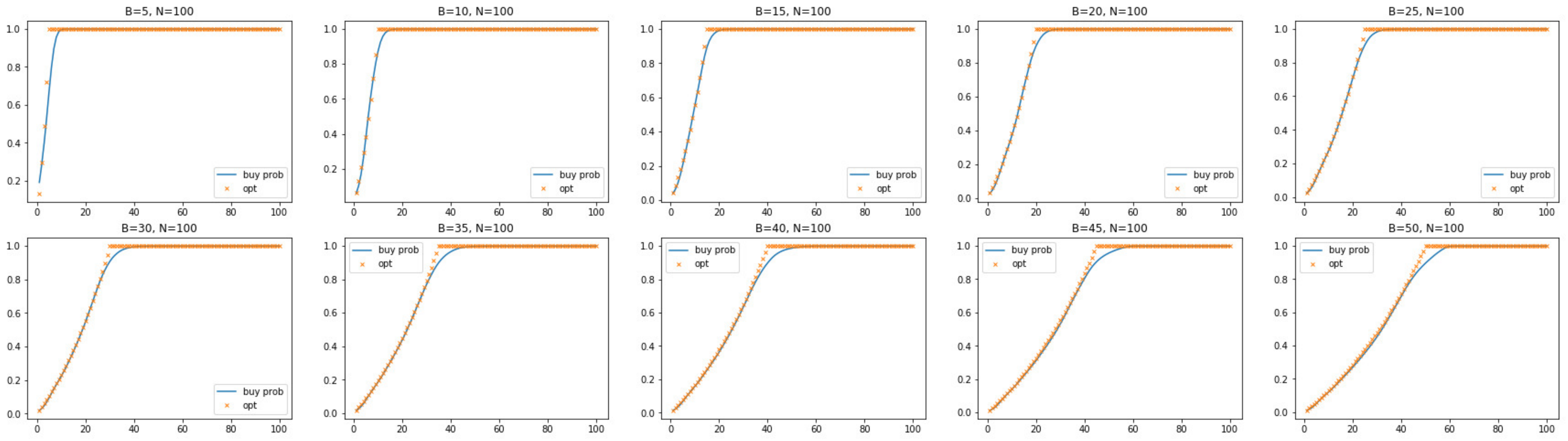}
    \caption{The learnt algorithm instantiated at various $(B,N)$ values as CDF of buying over time (blue curve). The x-axis denotes the number of days the algorithm has seen so far and the y-axis denotes the buying probability. The blue curve is the trained agent's output, while the orange correspond to the theoretical optimal strategy.}
    \label{fig:continuousskirental}
  \end{center}
\end{figure}

We demonstrate the trained algorithm in Figure~\ref{fig:continuousskirental} by instantiating it at various values of $B$ and $N$. The result is very close to the optimal strategy in all cases.
We also note here that the competitive ratio achieved by this algorithm converges approximately to $1.59$ within a few minutes of training, about $0.01$ worse than the optimal value ($1.582$).



\section{AdWords}\label{sec:appendix-adwords}

\subsection{Notable input distributions for AdWords}\label{sec:app-adwords-inputs}

\begin{figure}
	\centering
	\begin{subfigure}{.24\textwidth}
  		\centering
  		\includegraphics[width=0.5\textwidth,height=3cm]{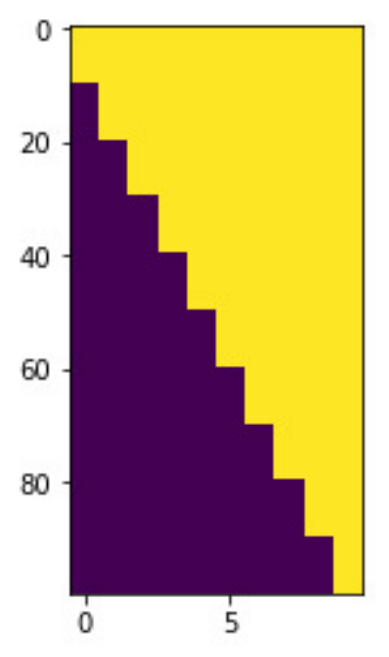}
  		\caption{The triangular graph}
  		\label{fig:triangular-graph}
	\end{subfigure}%
	\begin{subfigure}{.24\textwidth}
  		\centering
  		\includegraphics[width=0.5\textwidth,height=3cm]{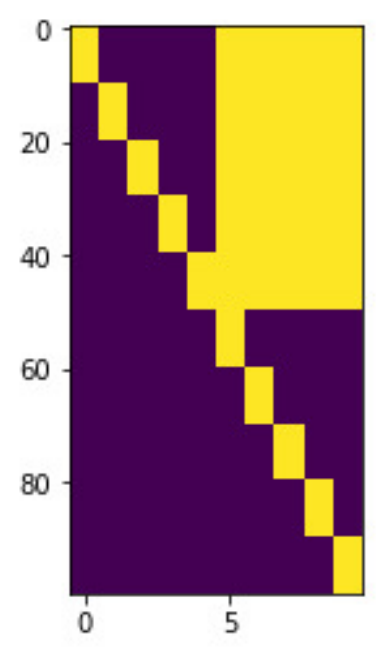}
  		\caption{The thick-z graph}
  		\label{fig:thick-z-graph}
	\end{subfigure}
	\caption{A graphical representation of the triangular and thick-z graphs as matrices. Ads are represented by rows and arrive from top to bottom, advertisers are represented by columns. A specific (ad, advertiser) cell represents the bid value where bright/dark cells indicate $1/0$. The graphs shown are of size $m\times n = 100\times 10$, and the same patterns generalize to larger instances.} 
\label{fig:graphs}
\end{figure}

\paragraph{Triangular distribution (\Cref{fig:triangular-graph}).} The triangular distribution certifies no algorithm for AdWords in the worst-case can get better than a CR of $1 - 1/e$ asymptotically (i.e., for large enough $m$ and $n$). The triangular distribution is based on the triangular matrix\footnote{Can be equivalently viewed as a bipartite graph where edges connecting ads and advertisers with edge weights representing the bid values ($0$ bid if there is no edge between an ad and an advertiser).} with the following pattern. There are $B n$ ads, $n$ advertisers, and each advertiser has a budget of $B$. An advertiser $0 \le j  < n$ bids $1$ for each of the first $(j+1) \cdot B$ arriving ads, and $0$ for the later ads. We permute the columns (i.e., advertiser ids) of the triangular matrix so the ids won't reveal the future, and the triangular distribution is the uniform distribution over all the column permuted matrices of the triangular matrix. The optimal offline solution has value $Bn$ by picking the block diagonal cells, while any algorithm (on a sufficiently large instance) cannot achieve better than $1 - 1/e$. The intuition behind this is that online algorithms has to guess and first allocate ads to the advertiser who is going to ``drop out'' early, which is impossible to do consistently when the columns are shuffled. The optimal allocation strategy is to make the spend of all advertisers as even as possible, i.e., Balance. See \cite{MSVV07} for details.

\paragraph{Thick-z distribution (\Cref{fig:thick-z-graph}).} The Thick-z distribution is an example for which the Greedy strategy under-performs compared to Balance. Suppose $n$ is even and let $B$ be an integer. There are $B n$ ads, $n$ advertisers, and each advertiser has a budget of $B$. An advertiser $0 \le i < n$ has bids $1$ with ads $\{iB, \ldots, (i+1)B - 1\}$ and, if $i \ge n/2$, also with $\{0, \ldots, Bn/2 - 1\}$. All other bids are $0$. The columns are then uniformly randomly permuted (to one of $n!$ permutations). The optimal offline solution has value $Bn$ again by picking the block diagonal cells in the thick-z graph. On the contrary, if some strategy (for example Greedy with randomized tie-breaking) is not smart at choosing the allocation for the earlier ads, it may allocate most of the first half of ads to the right half of the advertisers (in the unpermuted thick-z graph) and deplete their budget. When the second half of ads arrive, the advertisers with high bids for these ads already have their budget depleted, while those with remaining budget (i.e., the left half of advertisers) don't bid for the later ads.

\paragraph{Powerlaw distribution.} This distribution is inspired by high-level characteristics of real-world instances,
generated by a preferential attachment process on $m = n^2 \times n$ sized bipartite graphs. In particular, we refer to the $n$ advertisers as offline nodes and the $m$ ads as online nodes. For each online node $i$, we sample its degree from a log-normal distribution, that is, first sample $g_i \sim \mathrm{Gaussian}(1,1)$ and computes its degree as $d_i = \min(\exp(g_i), n)$. We then sample $d_i$ offline nodes (i.e., advertisers) to connect to the online node $i$ (i.e., bid non-zero for ad $i$), and the sampling probability of each offline node $j$ is proportional to the number of online nodes already connected to $j$. This random process generates instances where the degrees of offline nodes follow a powerlaw distribution, i.e., the long tail, which agrees with the observation of online advertising. The weights of these $d_i$ edges (i.e., the non-zero bids to ad $i$) are generated by first sampling its underlying value $v_i\sim \mathrm{Uniform}(0,1)$, and the individual bids are sampled i.i.d from the distribution $\mathrm{Gaussian}(v_i,0.1)$. 

As to the budgets of the advertisers, they are set non-uniformly in a way that makes the Greedy obtain the offline optimum. More specifically, we first run Greedy without budgets on an instance and simply let the result spending of each advertiser be the advertiser's budget, so running Greedy under these generated budgets achieve the best possible. This again agrees with the observation in online advertising where big advertisers bid high on many queries and have large budget, and that Greedy can exploit such typical structure to achieve very good performances. 

\paragraph{Triangular-g distribution.} Instances from this distribution are based on the $n^2\times n$ sized triangular graph. The differences between this distribution and the triangular distribution are as follows. First the bids are fractional instead of $0,1$, and each bid of $1$ in the triangular distribution is replaced by a random bid drawn i.i.d from $\mathrm{Uniform}(0.5,1)$ while bids of $0$ remain $0$. Furthermore, the budgets are no longer uniform in each instance and are computed the same way as in the Powerlaw distribution above.

\subsection{The algorithm learned by \YL~ discovers the salient features of MSVV}\label{sec:app-adwords}
In this section we demonstrate that \YL~ training successfully reconstructs the salient features of textbook-optimal worst-case algorithms and gives samples from the experience array. 

\begin{figure}[h]
  \begin{center}  
    \includegraphics[width=\textwidth]{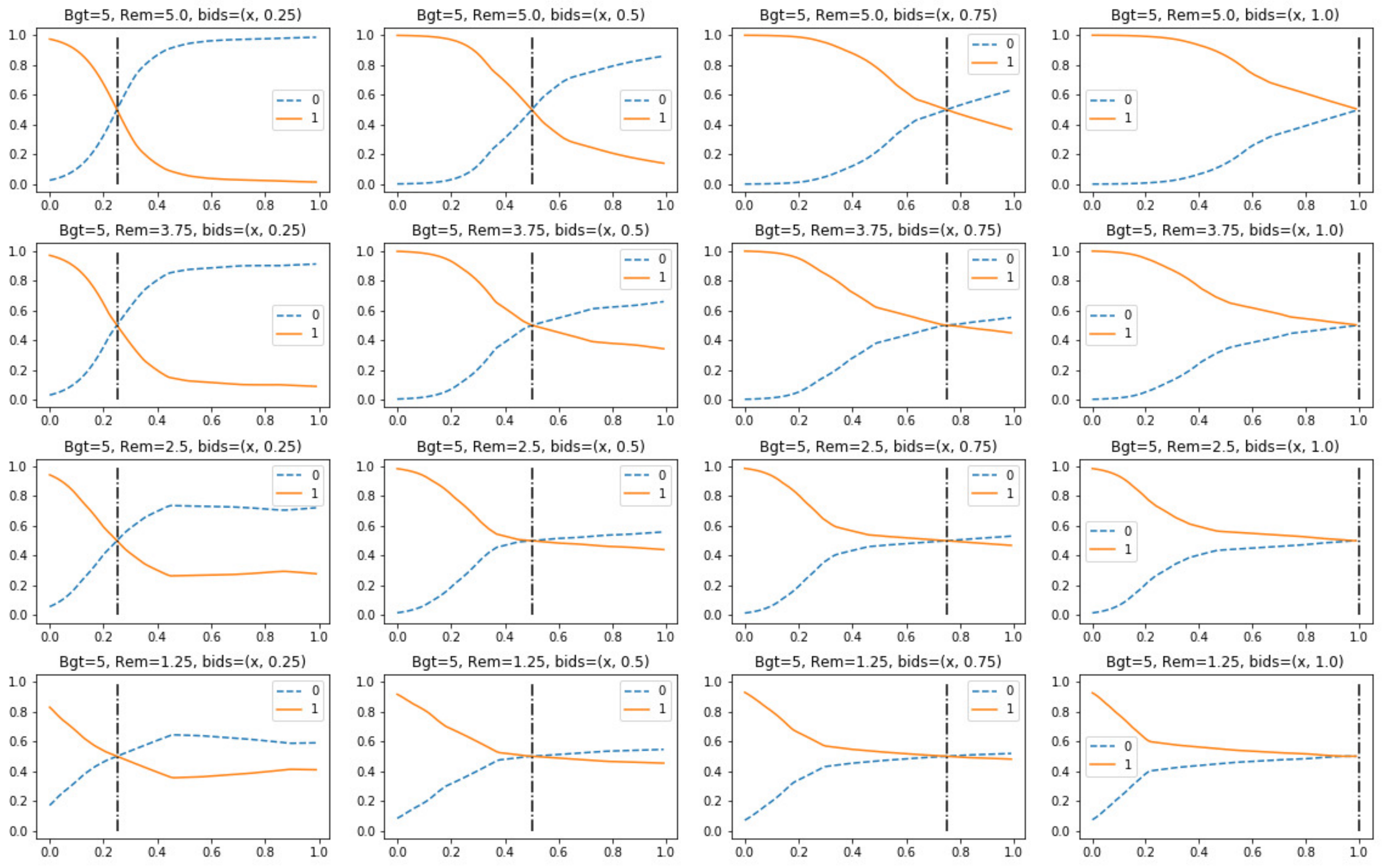}
    \caption{Plots demonstrating Greedy aspect of \YL~ learned algorithm.}
    \label{fig:adwords-weights}
  \end{center}
\end{figure}
\begin{figure}
  \begin{center}  
    \includegraphics[width=\textwidth]{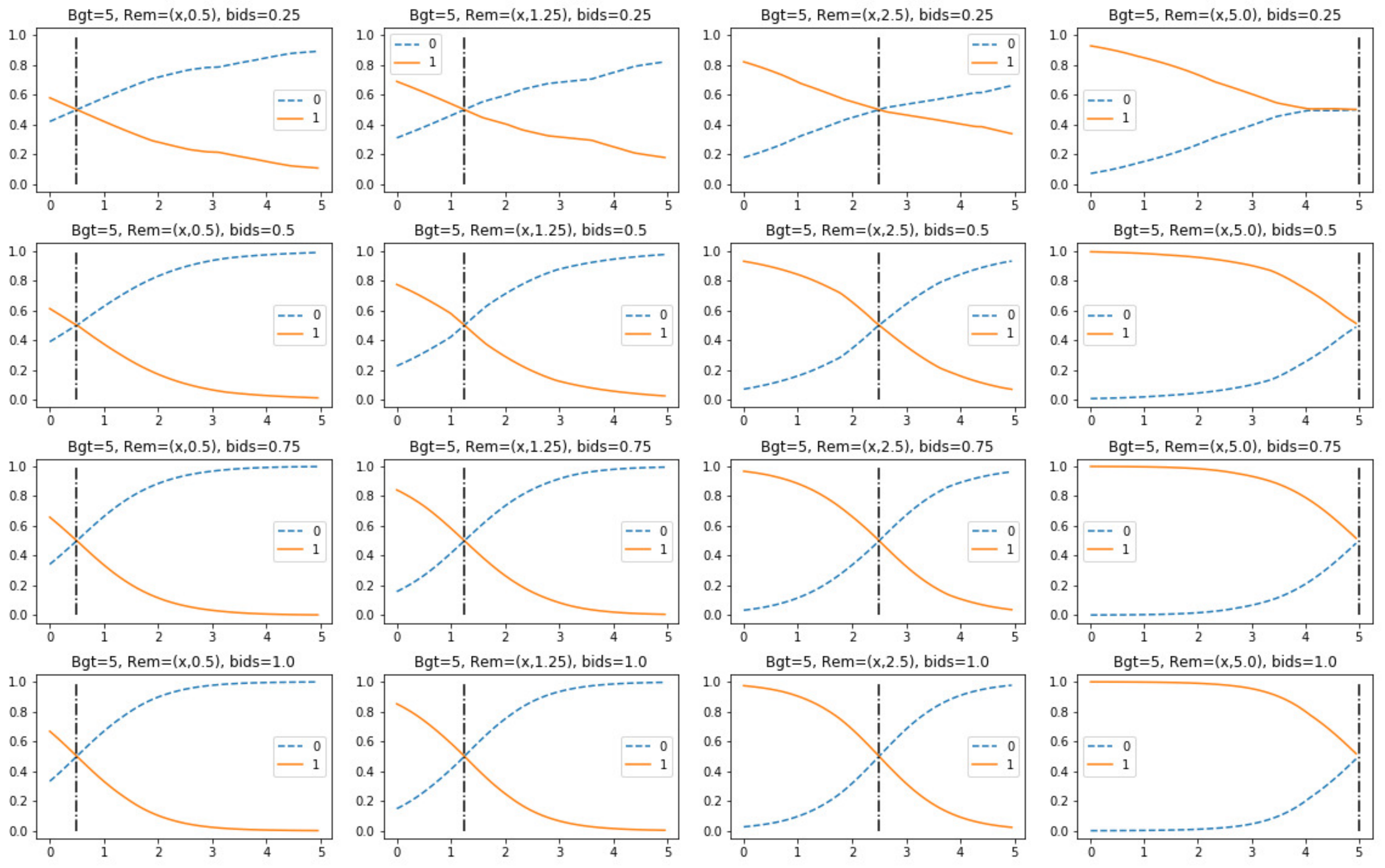}
    \caption{Plots demonstrating Balance aspect of \YL~ learned algorithm.}
    \label{fig:adwords-budgets}
  \end{center}
\end{figure}

\paragraph{Algorithm demonstrates both Greedy and Balance behavior.} 
The result in this section is using the algorithm trained on $25\times 5$ instances. In Figure~\ref{fig:adwords-weights} and Figure~\ref{fig:adwords-budgets} we interactively examine the behavior of the learned algorithm upon the arrival of one single ad slot with bids and fractional remaining budget inputs constructed specifically to test the Greedy and Balance aspects of the learned algorithm. In both figures, we have two advertisers $0,1$ and the vertical axis is the probability of assigning the ad slot to the advertisers respectively. In Figure~\ref{fig:adwords-weights}, we fix the fractional remaining budget of both advertisers at various levels ($0.25,0.5,0.75,1.0$). In each case, we fix the bid of advertiser $1$ and vary the bid of advertiser $0$. The optimal strategy should assign the ad to the advertiser with the highest bid since they all have same budget remaining. Our algorithm clearly demonstrates the correct \emph{greedy behavior} of the optimal algorithm in this scenario. The probability that advertiser $0$ receives the ad increases with his/her bid, and the value it takes for advertiser $0$ to be the most favorable (i.e., highest probability) matches fairly accurately the threshold where his/her bid becomes the highest. In Figure~\ref{fig:adwords-budgets}, we test whether the learned algorithm behaves accurately regarding the Balance aspect of the optimal algorithm. That is, when the bids are the same among both advertisers, the one with the highest remaining budget should win the ad. We fix the bid of both advertisers for the ad slot at various levels ($0.25,0.5,0.75,1.0$), fix the fractional remaining budget of advertiser $1$, and vary the remaining budget of advertiser $0$. Our algorithm clearly demonstrates the \emph{balance behavior} of the optimal algorithm fairly accurately.


\begin{figure}
	\centering
	\begin{subfigure}{.5\textwidth}
  		\centering
  		\includegraphics[width=0.5\textwidth,height=3cm]{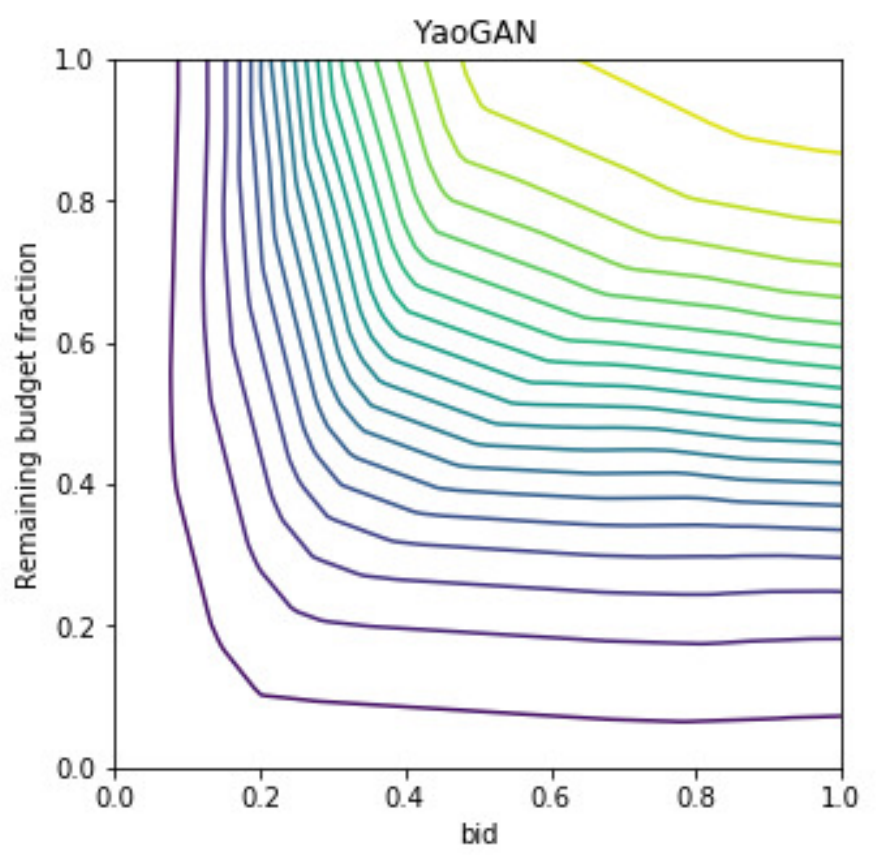}
  		\caption{tabula-rasa \YL~}
  		\label{fig:adv100-contour}
	\end{subfigure}%
	\begin{subfigure}{.5\textwidth}
  		\centering
  		\includegraphics[width=0.5\textwidth,height=3cm]{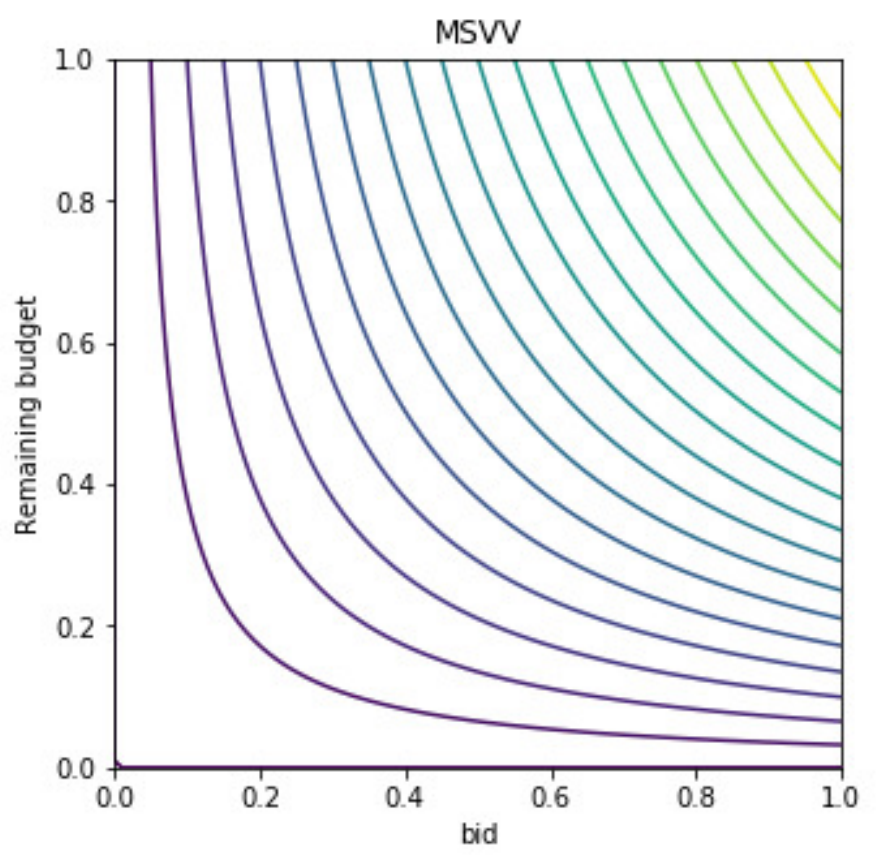}
  		\caption{MSVV}
  		\label{fig:msvv-contour}
	\end{subfigure}
	\begin{subfigure}{.5\textwidth}
  		\centering
  		\includegraphics[width=0.5\textwidth,height=3cm]{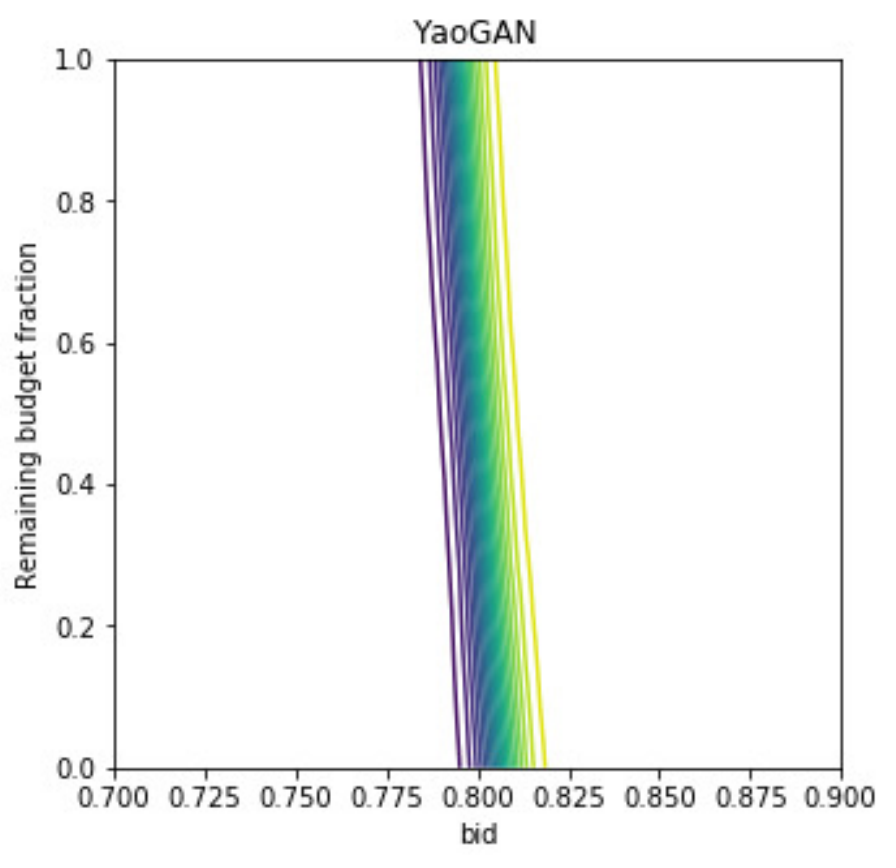}
  		\caption{pwl-100}
  		\label{fig:pwl100-contour}
	\end{subfigure}%
	\begin{subfigure}{.5\textwidth}
  		\centering
  		\includegraphics[width=0.5\textwidth,height=3cm]{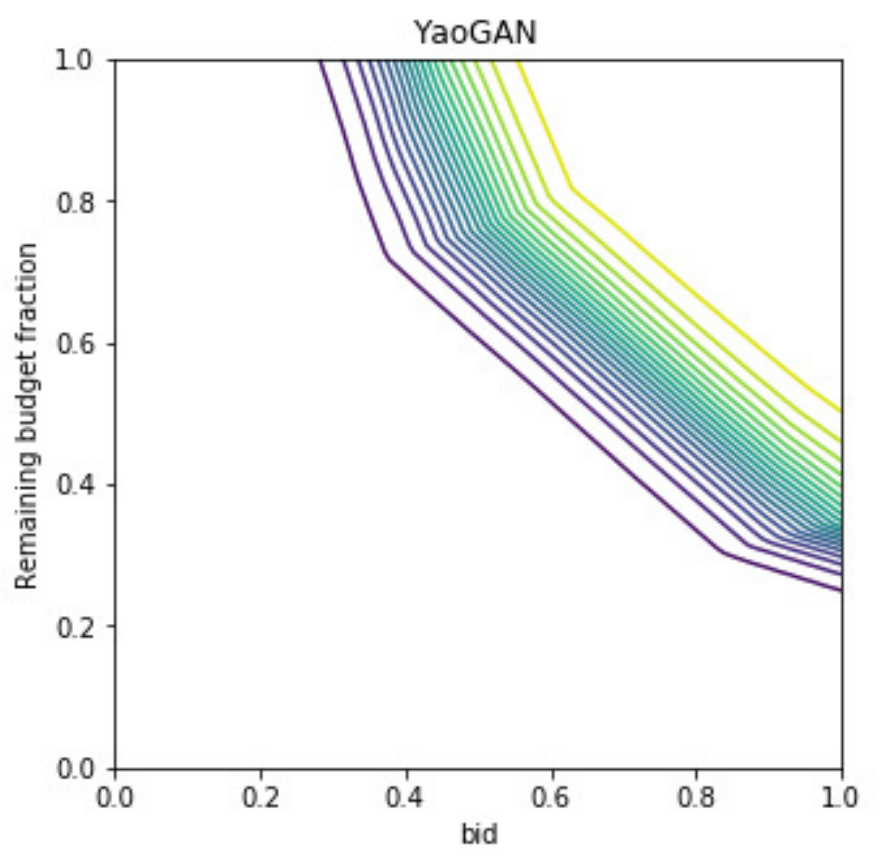}
  		\caption{pwl-90}
  		\label{fig:pwl90-contour}
	\end{subfigure}
	\caption{The contour plots of (a) \YL~ trained algorithm (tabula-rasa), (b) MSVV, (c) pwl-100, (d) pwl-90 demonstrating how the algorithms trade-off greedy and balance. In each plot, all the (bid, fractional remaining budget) pairs on the same contour curve are equivalent in terms of getting the ad.} 
\label{fig:contour-graphs}
\end{figure}

\paragraph{Algorithm demonstrates trade-off between Greedy and Balance.} We further probe the trained algorithm to see if it learns a good trade-off between Greedy and Balance. For this, we consider a scenario of two advertisers, where we vary both the bid and fractional remaining budget of advertiser $1$ while fixing the the bid and fractional remaining budget of advertiser $2$ to be $0.8$ and $0.5$ respectively, and both advertisers have total budget $5$. We consider all the $(x,y)$ pairs where $x,y$ are multiples of $1/100$ in $[0,1]$, i.e., we consider $100\times 100$ grid points over the $2$-dimensional space of $[0,1]\times[0,1]$. For each such $(x,y)$ pair, we set the bid and fractional remaining budget of advertiser $1$ to be $x$ and $y$ respectively, and run the \YL~ trained algorithm on this ad and two advertisers to get the probability of allocating the ad to advertiser $1$ which we refer to as $Z(x,y)$. In Figure~\ref{fig:adv100-contour} draw the contour plot on the $2$-dimensional space of $x$ and $y$ based on the $Z(x,y)$ value, so that all the pairs on the same contour lead to the same probability of advertiser $1$ getting the ad. The shape of the contour curves tell us in this particular scenario how the algorithm trade-off between bid value and remaining budget, i.e., the Greedy behavior and the Balance behavior. 

To provide some benchmark, we also consider how MSVV trade-off between bid value and remaining budget in general. For any $(x,y)$ pair representing the bid and fractional remaining budget of an advertiser for an ad, MSVV computes the scaled bid of $x \cdot (1-e^{-y})$ and pick the advertiser with the highest scaled bid. Thus we know all the $(x,y)$ pairs leading to the same scaled bid will be considered as equivalent to MSVV. In Figure~\ref{fig:msvv-contour} we draw the contour plot on the $2$-dimensional space of $x$ and $y$ based on the scaled bid value. The shape of the \YL~ trained algorithm's contour curves is very similar to the MSVV contour curves qualitatively. On the other hand, consider the simple strategy of Greedy, since only the bid value matters for Greedy, if we draw the contour plot for Greedy where the same $(x,y)$ pairs are equivalent to Greedy, the curves would all be vertical lines. 

We also draw the same contour plots for pwl-100 and pwl-90 (the algorithms trained by \YL~ in the robust stochastic setup) in Figure~\ref{fig:contour-graphs}. For pwl-100 the curves are very similar to Greedy (i.e., vertical lines) and we need to zoom the $x$-axis to be $[0.7,0.9]$ to see the slight difference because the other advertiser's bid is fixed at $0.8$ so bids much lower (or higher) than $0.8$ would give allocation probability of almost $0$ (or $1$) to advertiser $0$. We can see that while pwl-100 behaves much like Greedy in this context, pwl-90 demonstrates more of the Balance behavior as the fully adversarial trained algorithm.

\begin{figure}[h]
  \begin{center}  
    \includegraphics[width=\textwidth]{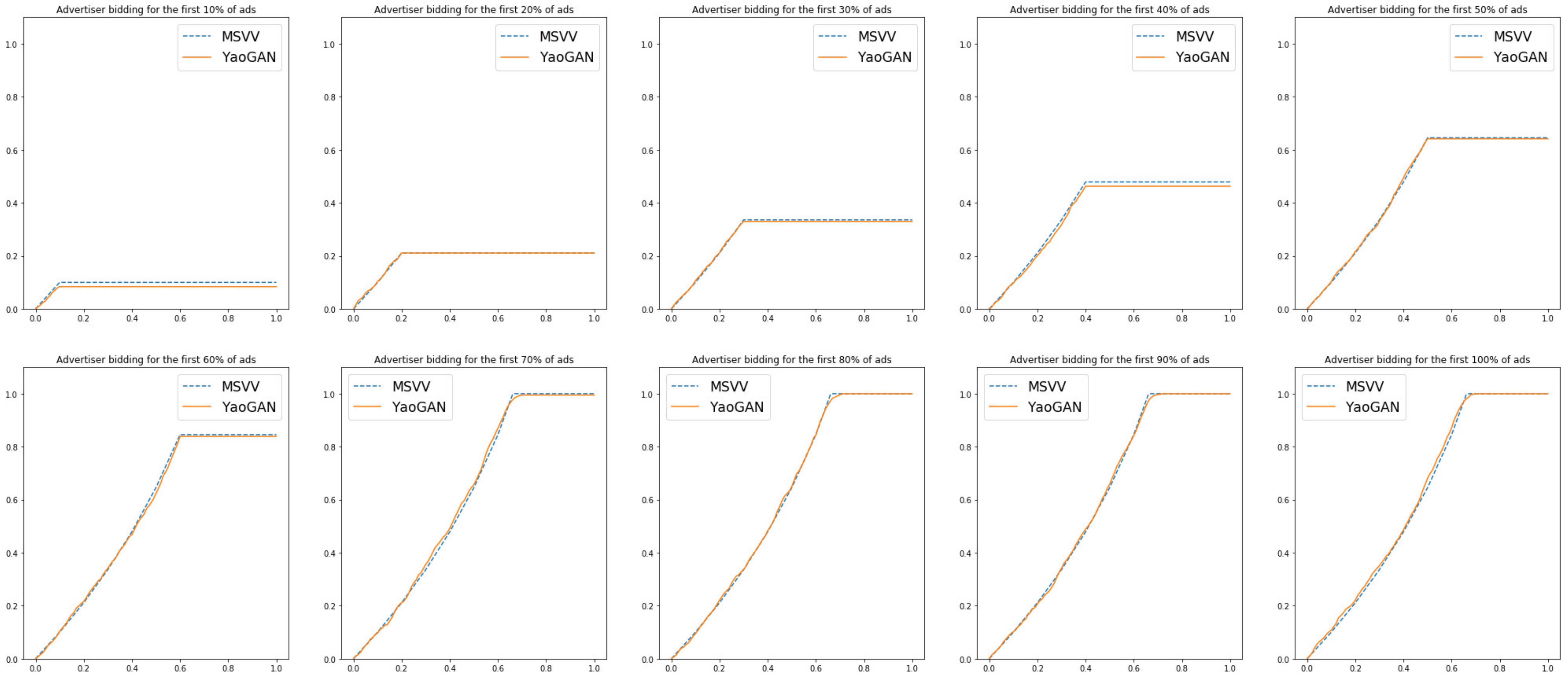}
    \caption{The evolution of the advertisers’ spending by \YL~ trained algorithm and MSVV for the (unpermuted) $100\times 10$ triangular matrix instance. Blue dotted curve is the expected spending of each advertiser if we run MSVV, and orange solid curve is the empirical expected spending over $100$ runs of \YL~ trained algorithm.}
    \label{fig:spend-triangular}
  \end{center}
\end{figure}

\begin{figure}[h]
  \begin{center}  
    \includegraphics[width=\textwidth]{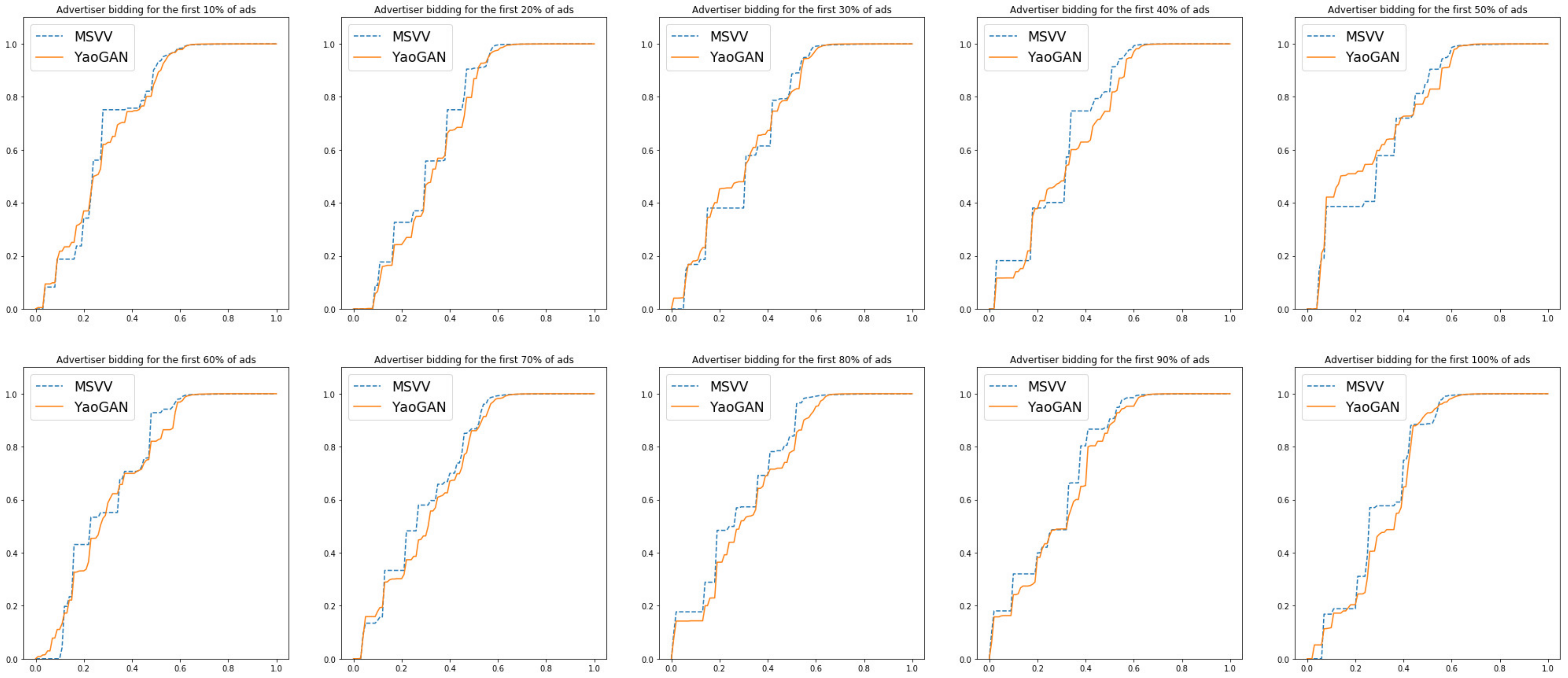}
    \caption{The evolution of the advertisers’ spending by \YL~ trained algorithm and MSVV on an randomly generated instance. Blue dotted curve is the empirical expected spending of each advertiser if we run MSVV, and orange solid curve is the empirical expected spending of \YL~ trained algorithm, both averaged over $100$ runs of the algorithms.}
    \label{fig:spend-random}
  \end{center}
\end{figure}

\paragraph{Detailed algorithm behavior on specific instances.} We further compare the spending behavior of the \YL~ trained algorithm and MSVV on three particular examples. We observe that the spending observed by \YL~ and MSVV are almost identical, which gives stronger evidence that \YL~ reconstructs the optimal algorithm than simply comparing the CR.
The first example is the $100\times 10$ triangular graph, where we have $10$ advertisers $0,\ldots,9$. Advertiser $j$ bids $1$ for the first $10(j+1)$ ads and bid $0$ for the rest. In Figure~\ref{fig:spend-triangular} we plot how the spending evolves for each advertiser as ads arrive. The $x$-axis indicates what fraction of the entire input has been revealed, and the $y$-axis is the fraction of total budget already spent. The $10$ subplots correspond to the $10$ advertisers in order. We also plot the spending by MSVV on this example, and our learned algorithm has nearly identical behavior as MSVV very closely on this example. As a second example, in Figure~\ref{fig:spend-triangular} we have the same plot for the example of $100\time 10$ thick-z graph. Again our algorithm is nearly identical to MSVV on this example. In Figure~\ref{fig:spend-random} we present an additional $100\time 10$ example where all the (weighted) bids are generated i.i.d from the uniform distribution over $[0,1]$ and the budget is $5$ for all advertisers. 

\begin{figure}
  \begin{center}  
    \includegraphics[width=\textwidth]{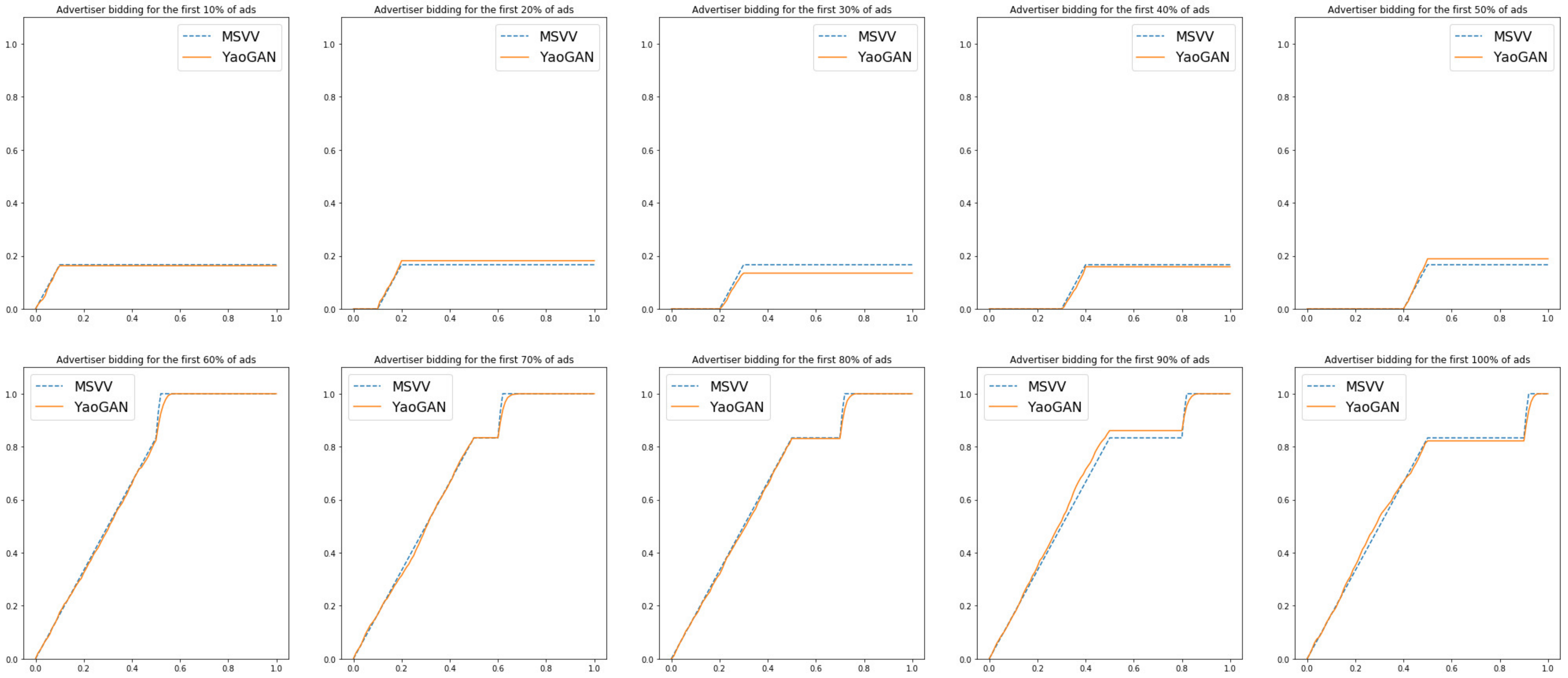}
    \caption{The evolution of the advertisers’ spending by \YL~ trained algorithm and MSVV for the (unpermuted) $100\times 10$ thick-z matrix instance. Blue dotted curve is the expected spending of each advertiser if we run MSVV, and orange solid curve is the empirical expected spending over $100$ runs of \YL~ trained algorithm.}
    \label{fig:spend-thickz}
  \end{center}
\end{figure}

\begin{figure}[h]
  \begin{center}  
    \includegraphics[width=0.8\textwidth,height=12cm]{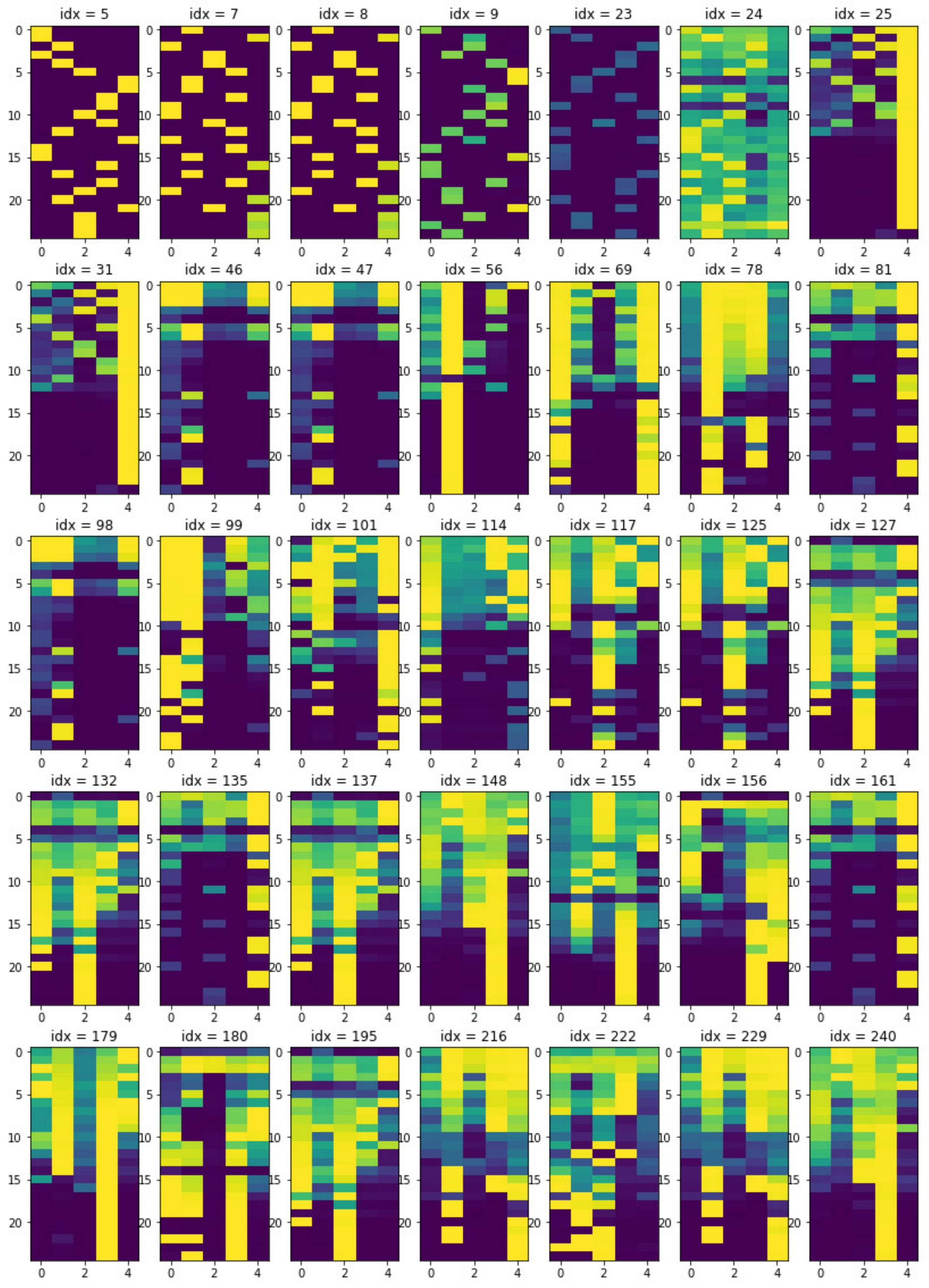}
    \caption{A set of random samples from the experience array after training. The $j^{th}$ row and $i^{th}$ column represent the bid of advertiser $i$ for ad slot $j$. Darker color represents a bid close to $0$, while brighter color represents a bid close to $1$.}
    \label{fig:adv}
  \end{center}
\end{figure}

\paragraph{Example instances generated by adversary network during \YL~ training.} We present some instances sampled from the experience array at the end of \YL~ training in Figure~\ref{fig:adv}. Recall these are instances with $5$ advertisers and $25$ ads and we only plot the bids of these instances without the advertisers' budgets (which are also generated by the adversary network) for easier visualization. In each matrix, ad slots are rows with the top row arriving first, brighter entries mean higher bids (between $0$ and $1$), and the matrices are ordered by when they are used to train the algorithm. Qualitatively, the first row depicts instances that teach the algorithm to reconstruct the Greedy strategy, while the rest resemble more difficulty instances that teach the algorithm the Balance strategy. These more difficulty instances have the following pattern: 1) many advertisers have high bids for the earlier ad slots, and 2) fewer advertisers have high bids for the ad slots arriving later. For such input, if the algorithm is not smart at choosing the allocation for the earlier ads, it may deplete the budget of some advertisers, and run into the situation where advertisers with high bids for the later ads already have their budget depleted, while those with remaining budget don't value highly of the later ads. Both the adversarial graph and thick-z graph have such structure. Although we don't have any quantitative measure, it is fairly clear that the pattern discussed above emerges as the training evolves. In particular, most of the instances generated later in the training clearly demonstrate the pattern that earlier ads (i.e., top rows) have more advertisers bidding highly (i.e., brighter cells), and later ads have less. It is very interesting that the adversarial neural network can come up with such pattern from scratch via interactively playing against the algorithm network.

\subsection{Training the adversary network against fixed algorithms}\label{sec:adv-training-fixed}
In this section we isolate the training of the adversary network against a fixed algorithm and show that the adversary network can find bad examples for a fixed algorithm fairly fast. We consider two fixed algorithms, MSVV and Greedy, since we know these algorithms and can understand why they perform poorly or well on particular examples. For easier understanding the examples presented in this section, we let the adversary network to only generate the bids of $25\times 5$ examples while fixing the budget of every advertiser to be $5$. 

\begin{figure}[H]
	\centering
	\begin{subfigure}{.5\textwidth}
  		\centering
  		\includegraphics[width=0.5\textwidth,height=3cm]{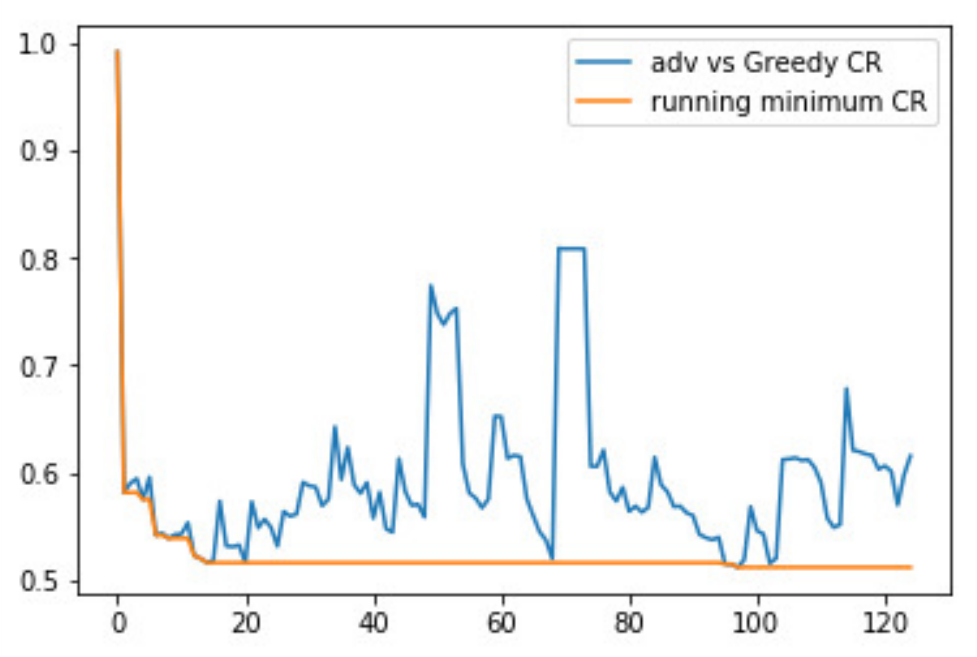}
  		\label{fig:adv-cr-greedy}
	\end{subfigure}%
	\begin{subfigure}{.5\textwidth}
  		\centering
  		\includegraphics[width=0.5\textwidth,height=3cm]{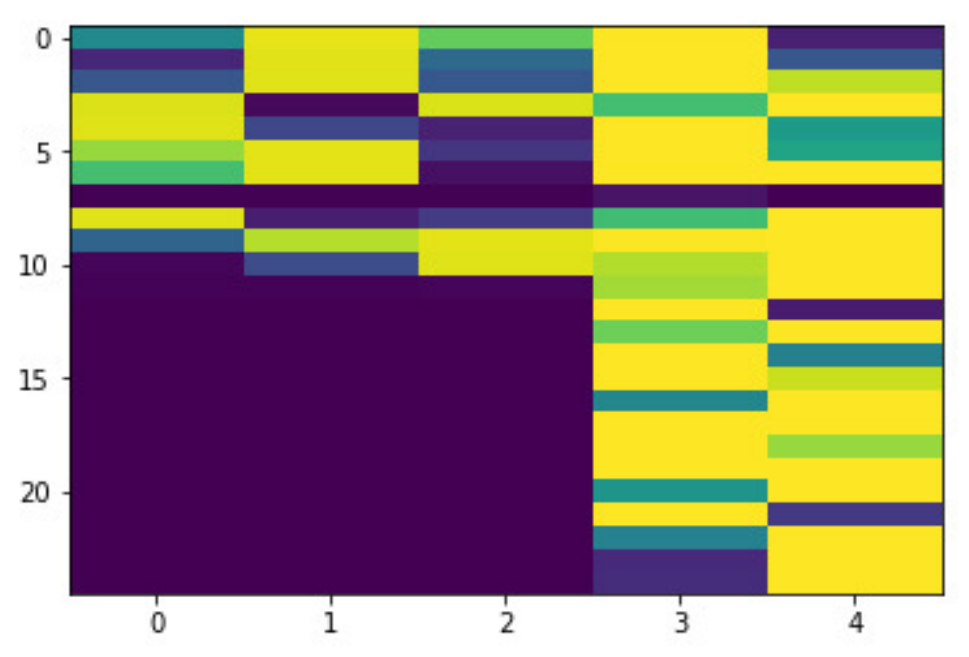}
  		\label{fig:adv-greedy-input}
	\end{subfigure}
	\caption{The CR of adversary generated examples against Greedy and the worst instance found.} 
\label{fig:adversary-greedy}
\end{figure}

\begin{figure}[H]
	\centering
	\begin{subfigure}{.5\textwidth}
  		\centering
  		\includegraphics[width=0.5\textwidth,height=3cm]{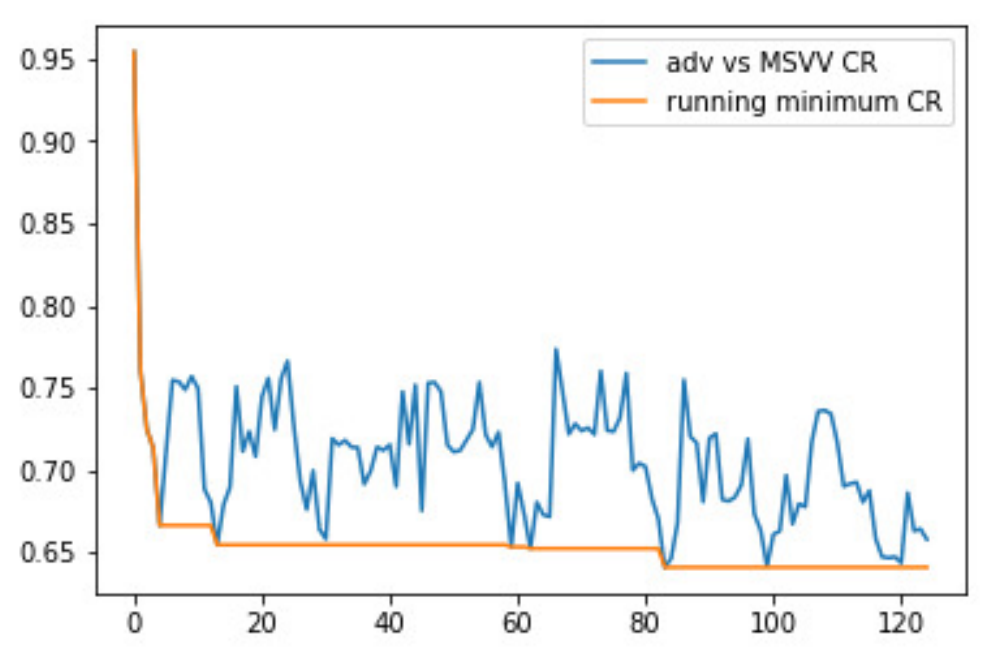}
  		\label{fig:adv-cr-msvv}
	\end{subfigure}%
	\begin{subfigure}{.5\textwidth}
  		\centering
  		\includegraphics[width=0.5\textwidth,height=3cm]{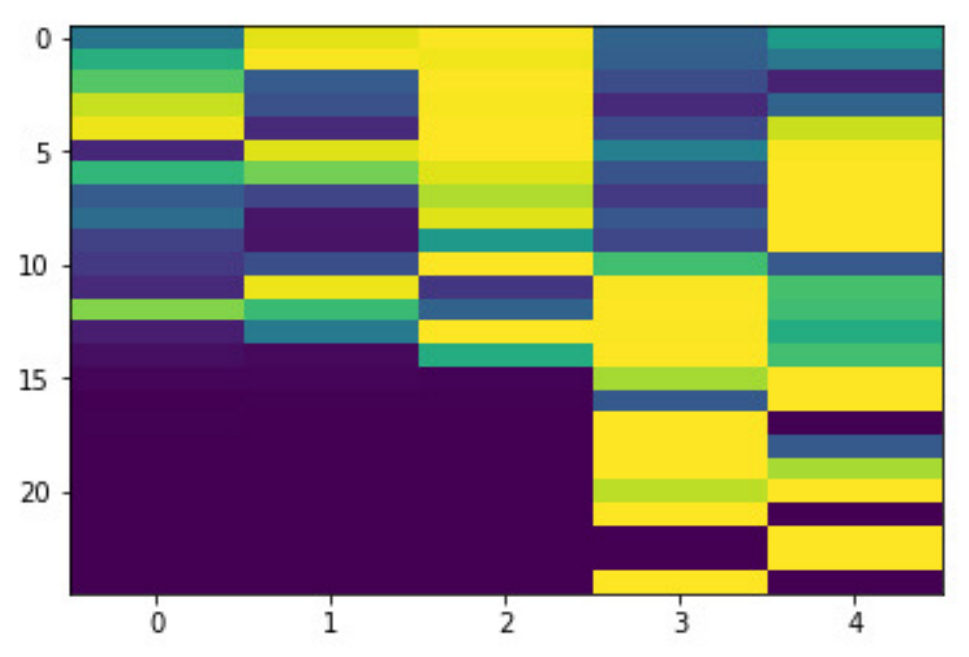}
  		\label{fig:adv-msvv-input}
	\end{subfigure}
	\caption{The CR of adversary generated examples against MSVV and the worst instance found.} 
\label{fig:adversary-msvv}
\end{figure}

In Figure~\ref{fig:adversary-greedy} we present the result against Greedy. The left plot is the CR of Greedy on the generated examples as time evolves. We plot the CR for the last example generated by the adversary network in every $100$ training steps (the blue curve), and the orange curve is the running minimum CR of Greedy on previous examples. We also reinitialize the adversary network every $500$ training steps. Greedy has a worst-case CR of $0.5$, and We can see that the adversary network finds bad examples approach the worst-case CR for Greedy fairly fast. In the right plot we draw the worst instance generated by adversary for Greedy with a CR of $0.512$. It is not hard to see why the instance is bad for Greedy, as the offline optimal allocation is to allocate the first half of ads between the first three advertisers and save the budget of the right-most two advertisers for the later ads. On the contrary, since the highest bids of the first half of ads are all among the right-most two advertisers, Greedy will allocate these ads to them and they run out of budget before seeing the later ads. In Figure~\ref{fig:adversary-msvv} we present the analogous plots for adversary training against MSVV. The lowest CR of MSVV on adversary generated examples is $0.640$, which is very close to its worst-case CR of $1-1/e$.

\begin{figure}[H]
	\centering
	\begin{subfigure}{.5\textwidth}
  		\centering
  		\includegraphics[width=0.5\textwidth,height=3cm]{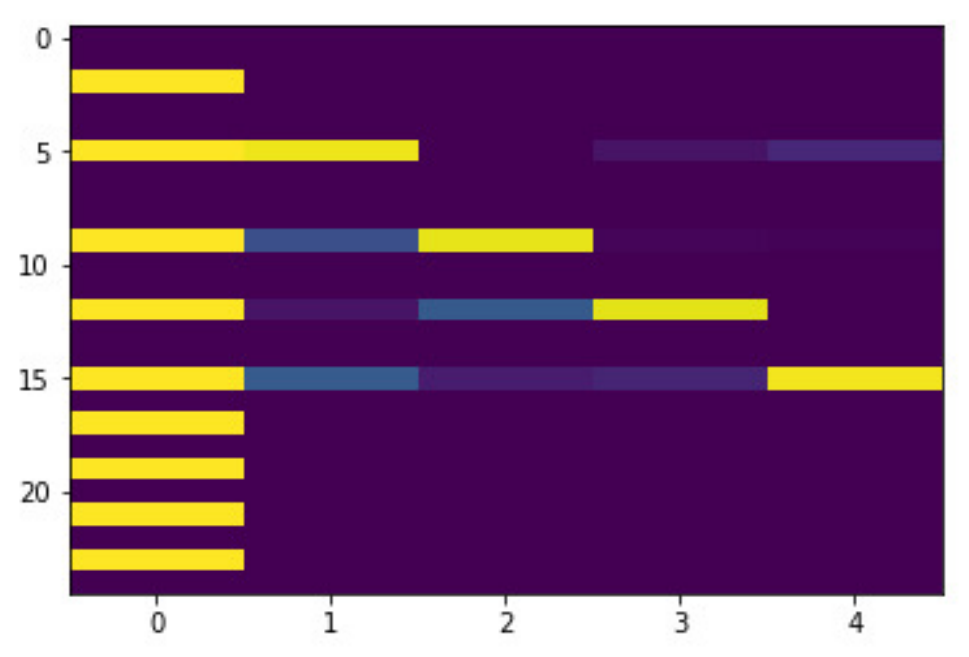}
  		\label{fig:msvv-greedy}
	\end{subfigure}%
	\begin{subfigure}{.5\textwidth}
  		\centering
  		\includegraphics[width=0.5\textwidth,height=3cm]{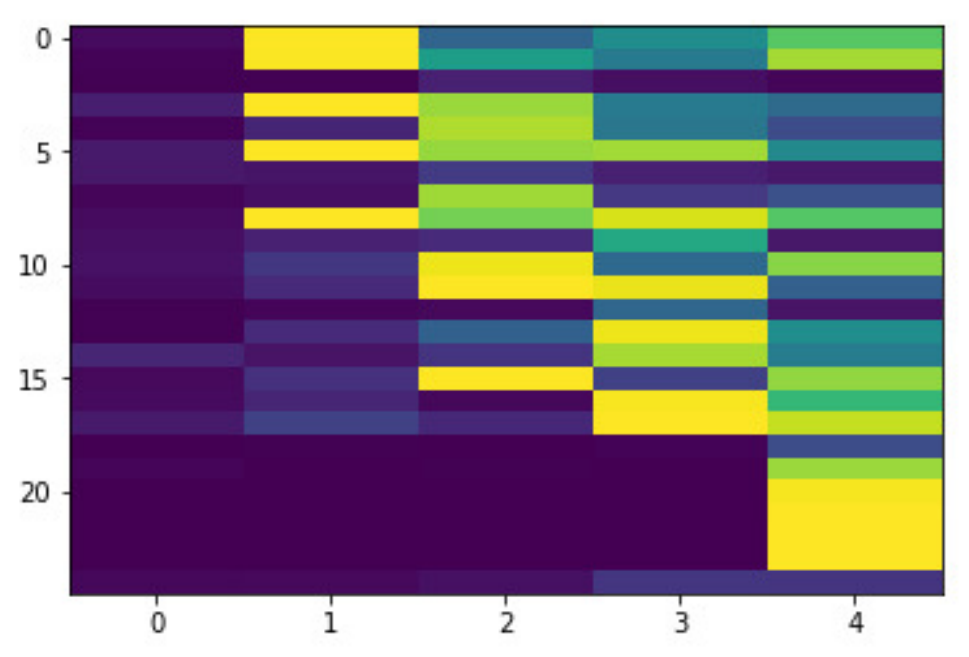}
  		\label{fig:greedy-msvv}
	\end{subfigure}
	\caption{The instances generated by the adversary network where MSVV outperforms Greedy by the most (left) and vice versa (right).} 
\label{fig:diff_examples}
\end{figure}

We note that beyond using the adversary network to generate difficult instances for a fixed algorithm as discussed above, we can easily modify the training objective to provide other intuitions that are valuable in understanding algorithms. For example, we can make the training loss of the adversary network be the difference between CRs of MSVV and Greedy on its output instances (instead of just the CR of one fixed algorithm), and use it to find instances where one algorithm out-performs another algorithm by the largest margin. In Figure~\ref{fig:diff_examples} we present the instances generated by the adversary network in this setting. The left plot is when we ask for an instance where MSVV outperforms Greedy, and the right plot is an instance where Greedy outperforms. Again it is very interesting that the adversary network can construct instances with such interesting structures from scratch. On the left example, the large bids of the first advertiser (the bright cells in the left-most column) are actually larger than the bright cells in the other columns by a tiny amount. Greedy on this example will only collect revenue on the first $5$ bright cells from the first advertiser and then that advertiser runs out of budget so we cannot collect revenue from the last $4$ bright cells of that advertiser. On the contrary, as the large bids have roughly the same value, MSVV will allocate the top bright cell to the first advertiser, and then for the subsequent $4$ ads where the first advertiser and another advertiser both bid high, MSVV will allocate to the other advertiser. Then for the last $4$ ads where the first advertiser bids high, there is still budget left (recall budget is $5$ for everyone) so we can allocate to that advertiser. The instance on the right is one where Greedy outperforms MSVV by a large margin. The structure of the instance is very similar to the triangular graph but weighted so the block diagonal cells have larger value in general. Greedy would behave very similarly to the offline optimal in allocating the block diagonal cells and get very good CR. On the contrary, MSVV would behave sub-optimally because of the Balance aspect of the algorithm.

\subsection{Permutation equivariant algorithm network}\label{sec:PE-networks} 
Recall our algorithm network takes as input $n$ triples (one for each advertiser) that represent the bid $v_i^j$ to the incoming ad slot $j$, the fractional remaining budget $r_i^j / B_i$, and the total budget $B_i$.  It computes, for each advertiser, the probability that the incoming ad slot will be allocated to that advertiser. To accomplish this, the algorithm network comprises a single-agent neural network that computes a score for a generic advertiser, and a soft-max layer that converts the scores into probability values. The single-agent network takes a $6$-dimensional input vector
consisting of the three quantities $v_i^j$, $r_i^j/B_i$, and $B_i$, and the respective sums of these three quantities over all advertisers. Upon receiving a $3n$-dimensional input for an ad, the algorithm network cuts the input into $n$ individual vectors each of size $6$, feed each such vector to the (shared) single-agent network. The algorithm network collects the $n$ scalar outputs from the single-agent network and takes a softmax over them to generate the output $n$-dimensional allocation probabilities. Note the number of trainable weights in our algorithm network does not grow with $n$.

The motivation of using such a neural network structure for the algorithm is two-fold. Firstly, it guarantees that the learned algorithm is \emph{permutation equivariant} (PE), which means if we permute the $n$ advertisers in the input, then the output of the algorithm should undergo the same permutation. In other words, the algorithm should make decisions not based on the ids of the advertisers. This is a desirable property for algorithms of AdWords and also the correct property as long as the ids don't correlate with any future information. Secondly, as the structure and number of trainable weights of the single-agent network aren't affected by the number of advertisers $n$, it is straightforward to extend the trained algorithm to work on instances when $n$ changes. This together with the online nature of problem makes our learned algorithm an {\em uniform} algorithm, i.e., it can work on arbitrary sized input.
We note that we can use more involved network structure if required while still guaranteeing that the learned algorithm is PE and uniform. For interested readers, we refer to the work of~\citep{ZaheerKRPSS17,HartfordGLR18} for more detailed discussion of enforcing PE in neural networks, and~\citep{Rahme20} for an application in auction, which is another canonical multi-agent setting where PE is desired. 


\bibliography{refs}
\bibliographystyle{plainnat}

\end{document}